\documentclass{article}

\usepackage[preprint, nonatbib]{neurips_2022}

\usepackage[utf8]{inputenc} % allow utf-8 input
\usepackage[T1]{fontenc}    % use 8-bit T1 fonts
\usepackage[pagebackref,breaklinks,colorlinks,citecolor=citecolor2]{hyperref}
\usepackage{url}            % simple URL typesetting
\usepackage{booktabs}       % professional-quality tables
\usepackage{amsfonts}       % blackboard math symbols
\usepackage{nicefrac}       % compact symbols for 1/2, etc.
\usepackage{microtype}      % microtypography
\usepackage{xcolor}         % colors

\usepackage{color}
\usepackage[pdftex]{graphicx}
\usepackage{array,multirow}
\usepackage{amsmath, mathtools, bm, bbm}
\usepackage{amsthm}
\usepackage{amssymb}
\usepackage{tabularx}
\usepackage{soul} % for underline that wraps, i.e., \ul
\usepackage{pifont}% http://ctan.org/pkg/pifont
\usepackage{subcaption}
\usepackage{lipsum}
\definecolor{Gray}{gray}{0.95}
\usepackage{colortbl} % color row in table

\usepackage{times}
\usepackage{epsfig}
\usepackage{makecell}
\usepackage{bbding}
\usepackage{caption}
\usepackage[inline]{enumitem} 
\usepackage{wrapfig}

\definecolor{codegreen}{rgb}{0.0,0.6,0.0}
\definecolor{darkergreen}{RGB}{21, 152, 56}
\definecolor{red2}{RGB}{252, 54, 65}
\definecolor{cadmiumgreen}{rgb}{0.0, 0.42, 0.24}
\definecolor{cadmiumred}{rgb}{0.89, 0.0, 0.13}
\definecolor{myblue}{rgb}{0.0, 0.15, 0.90}
\definecolor{LightCyan}{rgb}{0.88,1,1}
\newcommand{\green}[1]{\textcolor{cadmiumgreen}{#1}}

\newcommand{\blue}[1]{\textcolor{myblue}{#1}}

\newcommand{\sd}[1]{\textcolor{cyan}{#1}}
\newcolumntype{g}{>{\columncolor{LightCyan}}l}
\newcolumntype{h}{>{\columncolor{LightCyan}}c}
\newcommand{\ms}[1]{\tiny{$\pm$#1}}

\usepackage{listings}
\usepackage{algorithm}
\usepackage{algcompatible}
\algnewcommand\algorithmicreturn{\textbf{return}}
\algnewcommand\RETURN{\State \algorithmicreturn}%
\algnewcommand\algorithmicfunction{\textbf{function}}
\algdef{SE}[FUNCTION]{Function}{EndFunction}%
   [2]{\algorithmicfunction\ \text{#1}\ifthenelse{\equal{#2}{}}{}{(#2)}}%
   {\algorithmicend\ \algorithmicfunction}%
   
\definecolor{citecolor2}{HTML}{0071bc}

\title{Exploring Temporally Dynamic Data Augmentation for Video Recognition}

\author{
Taeoh Kim \\
NAVER Clova \\
\And
Jinhyung Kim \\
KAIST \\
\And
Minho Shim \\
NAVER Clova \\
\And
Sangdoo Yun\\
NAVER AI Lab \\
\And
Myunggu Kang \\
NAVER Clova \\
\And
Dongyoon Wee \\
NAVER Clova \\
\And
Sangyoun Lee \\
Yonsei University \\
}

\begin{document}

\maketitle

\begin{abstract}

Data augmentation has recently emerged as an essential component of modern training recipes for visual recognition tasks.
However, data augmentation for video recognition has been rarely explored despite its effectiveness.
Few existing augmentation recipes for video recognition naively extend the image augmentation methods by applying the same operations to the whole video frames.
Our main idea is that the magnitude of augmentation operations for each frame needs to be changed over time to capture the real-world video's temporal variations.
These variations should be generated as diverse as possible using fewer additional hyper-parameters during training.
Through this motivation, we propose a simple yet effective video data augmentation framework, DynaAugment.
The magnitude of augmentation operations on each frame is changed by an effective mechanism, Fourier Sampling that parameterizes diverse, smooth, and realistic temporal variations.
DynaAugment also includes an extended search space suitable for video for automatic data augmentation methods.
DynaAugment experimentally demonstrates that there are additional performance rooms to be improved from static augmentations on diverse video models.
Specifically, we show the effectiveness of DynaAugment on various video datasets and tasks: large-scale video recognition (Kinetics-400 and Something-Something-v2), small-scale video recognition (UCF-101 and HMDB-51), fine-grained video recognition (Diving-48 and FineGym), video action segmentation on Breakfast, video action localization on THUMOS'14, and video object detection on MOT17Det.
DynaAugment also enables video models to learn more generalized representation to improve the model robustness on the corrupted videos.
    
\end{abstract}
\section{Introduction}

Data augmentation is a crucial component of machine learning tasks as it prevents overfitting caused by a lack of training data and improves task performance without additional inference costs.
Many data augmentation methods have been proposed across a broad range of research fields, including image recognition~\cite{cubuk2019autoaugment, cubuk2020randaug, hendrycks2019augmix, devries2017improved, zhang2017mixup, yun2019cutmix, lingchen2020uniformaugment, muller2021trivialaugment}, image processing~\cite{yoo2020rethinking, yu2020bitmix}, language processing~\cite{sennrich2015improving, wei2019eda, wang2018switchout, chen2020mixtext}, and speech recognition~\cite{park2019specaugment,meng2021mixspeech}.
In image recognition, each augmentation algorithm has become an essential component of the modern training recipe through various combinations~\cite{touvron2021training, bello2021revisiting, wightman2021resnet, liu2022convnet}.
However, data augmentation for video recognition tasks has not been extensively studied yet beyond the direct adaptation of the image data augmentations.

\begin{figure}[!t]
	\centering
	\begin{subfigure}[c]{0.49\textwidth}
		\centering
		\includegraphics[width=\linewidth]{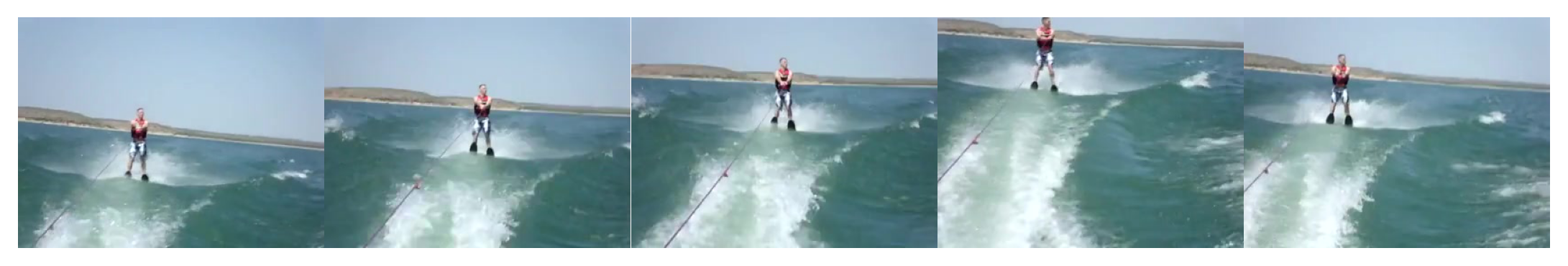}
	\end{subfigure}
	\hfill
	\begin{subfigure}[c]{0.49\linewidth}
    	\centering
    	\includegraphics[width=\linewidth]{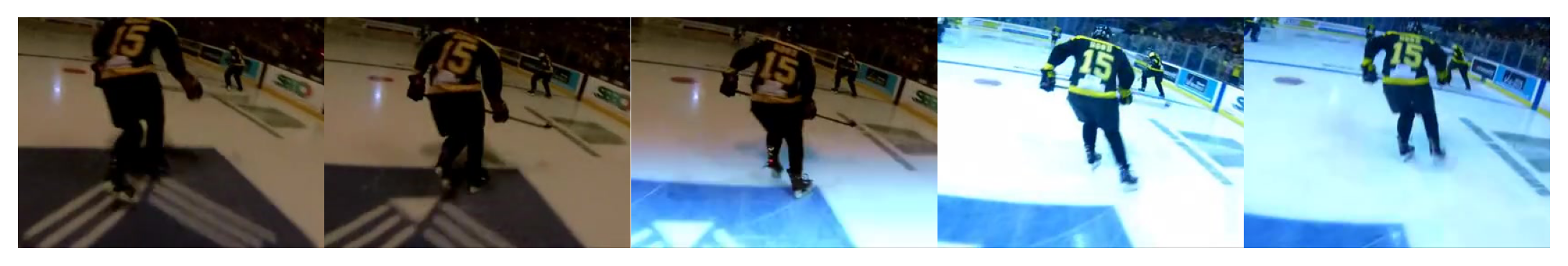}
	\end{subfigure}
	\\
\caption{
Applying static data augmentations has limitations in modeling temporally dynamic variations of real-world video.
Figure shows examples of temporal variations in videos.
(Left) \textit{geometric changes (water skiing)} and (Right) \textit{photometric changes (ice hockey)}. 
These variations are changed in an arbitrary but smooth manner.
}
	\label{fig:teaser}
\end{figure}

An effective data augmentation method is required to cover the comprehensive characteristics of data.
Videos are characterized by diverse, dynamic temporal variations, such as camera/object movement or photometric (color or brightness) changes.
As shown in Fig.~\ref{fig:teaser}, dynamic temporal changes make it difficult to find out the video's category for video recognition. 
Therefore, it is important to make the video models effectively cope with all possible temporal variations, and it would improve the model's generalization performance.

Recent achievements in video recognition research show notable performance improvements by extending image-based augmentation methods to video~\cite{yun2020videomix, qian2020spatiotemporal}.
Following image transformers~\cite{touvron2021training, touvron2021training,liu2021swin}, video transformers~\cite{fan2021multiscale, arnab2021vivit, liu2021video, li2022uniformer} have also been applied mixture of augmentations, but they are still static over video frames.
Their augmentations do not consider aforementioned temporal dynamics, simply applying the same augmentation over every frame for each video.
Even though applying well-studied image augmentation (\textit{e.g.} RandAugment~\cite{cubuk2020randaug}) statically on videos shows good results, there is still room for performance improvement via considering temporally dynamic factors.
From this motivation, an augmentation operation on each frame should be changed dynamically rather than being static over time.
To this end, we propose a simple yet effective data augmentation framework for video recognition called DynaAugment. 

However, the additional temporal axis of videos introduces new dimensions for controlling augmentations such as minimum and maximum magnitude, amplitudes of variation, the direction of magnitude changes, or frequencies of the fluctuation.
These increasing possibilities can cause extensive searching processes and computational costs to find the optimal augmentation policy.
A unified function is required to reduce the range of the parameterization as temporal variations (either geometrically or photometrically) in the real-world videos are generally linear, polynomial, or periodic (\textit{e.g.} moving objects, camera panning/tilting, or hand-held camera).
The key factor is changing the magnitude of augmentation operation as a function of time with a simple parameterization.
Based on Fourier analysis, an arbitrary signal can be decomposed into multiple basis functions.
All variations, as mentioned above, can be represented by the random weighted sum of diverse-frequency sinusoidal basis functions.
From this motivation, we propose a generalized sampling function called Fourier Sampling that generates temporally diverse and smooth variations as functions of time. 

DynaAugment also includes an extended search space suitable for video for automatic data augmentation methods.
Because existing state-of-the-art automatic augmentation methods such as RandAugment~\cite{cubuk2020randaug}, UniformAugment~\cite{lingchen2020uniformaugment}, and TrivialAugment~\cite{muller2021trivialaugment} inherit the augmentation operations from AutoAugment~\cite{cubuk2019autoaugment} that can be extended considering the video's characteristics.

To verify the effectiveness of the proposed method, we conduct extensive experiments on the video recognition task, where DynaAugment reaches a better performance than the static versions of state-of-the-art image augmentation algorithms.
The experimental results also demonstrate the generalization ability of DynaAugment.
Specifically, recognition performances are improved in both different types of models and different types of datasets including: large-scale dataset (Kinetics-400~\cite{carreira2017quo} and Something-Something-v2~\cite{goyal2017something}), small-scale dataset (UCF-101~\cite{soomro2012ucf101} and HMDB-51~\cite{kuehne2011hmdb}), and fine-grained dataset (Diving-48~\cite{li2018resound} and FineGym~\cite{shao2020finegym}). Furthermore, DynaAugment shows better transfer learning performance on the video action segmentation, localization, and object detection tasks.
We also evaluate our method on the corrupted videos as an out-of-distribution generalization, especially in the low-quality videos generated with a high video compression rate. 
Since training with DynaAugment learns the invariance of diverse temporal variations, it also outperforms other methods in corrupted videos.
\section{Related Work}

\paragraph{Video Recognition}

For video recognition, in the early days, 3D convolutional networks (CNNs)~\cite{ji20123d, tran2015learning} and two-stream~\cite{simonyan2014two} networks are proposed.
In \cite{carreira2017quo}, a method that utilizes the inflation of pretrained weights from the ImageNet~\cite{deng2009imagenet} is proposed.
After the development of the large-scale Kinetics~\cite{carreira2017quo} dataset, deep 3D residual networks~\cite{hara2018can} with attention modules~\cite{wang2018non}, and an efficient design of spatio-temporal networks, such as spatial-temporal decomposition~\cite{xie2018rethinking, tran2018closer} or channel separation~\cite{tran2019video, feichtenhofer2020x3d}, have been proposed. 
In~\cite{feichtenhofer2019slowfast}, a two-path CNN is proposed to capture different frame rates.
In~\cite{feichtenhofer2020x3d}, the X3D family networks are proposed via a progressive architecture search for efficient video recognition.
In another branch, \cite{lin2019tsm, wang2021tdn} design temporal modeling modules on top of the 2D-CNNs for efficient training and inference.
Recently, transformer networks have proven to be strong architectural designs for video recognition with~\cite{neimark2021video, bertasius2021space, arnab2021vivit} or without~\cite{fan2021multiscale} image transformer~\cite{dosovitskiy2020image} pre-training.
Video transformers with attention in the local window~\cite{liu2021video} and with convolutions~\cite{li2022uniformer} have shown state-of-the-art results.

\paragraph{Data Augmentation}

First, network-level augmentations (sometimes also called regularization) randomly remove~\cite{srivastava2014dropout, ghiasi2018dropblock, huang2016deep, larsson2016fractalnet} or perturb~\cite{gastaldi2017shake, yamada2019shakedrop, verma2019manifold, li2020feature, wang2021regularizing, kim2020regularization} in-network features or parameters.
For video recognition, 3D random mean scaling (RMS)~\cite{kim2020regularization} randomly changes the low-pass components of spatio-temporal features.

In image-level augmentations, after a simple data augmentation, such as a rotation, flip, crop~\cite{krizhevsky2012imagenet}, or scale jittering~\cite{simonyan2014very}, algorithms that randomly remove spatial regions~\cite{devries2017improved, singh2017hide, zhong2020random} to prevent overfitting to the most discriminative parts have been proposed.
To further improve the performance, a learning-based data augmentation technique called AutoAugment~\cite{cubuk2019autoaugment} is proposed and has demonstrated remarkable performances.
However, AutoAugment requires many GPU hours to find the optimal augmentation policy.
RandAugment~\cite{cubuk2020randaug} suggests that the reduction of the search space. % is more critical and practical.
It has produced results comparable with those of AutoAugment with only two searching parameters.
Recently, search-free augmentation strategies including UniformAugment~\cite{lingchen2020uniformaugment} and TrivialAugment~\cite{muller2021trivialaugment} have shown similar performances using very simple parameterizations.
They also extend the search space for augmentation operations that are inherited from those of AutoAugment to improve the performances.
MixUp~\cite{zhang2017mixup} and CutMix~\cite{yun2019cutmix} randomly mix the training samples; this results in both the strengths of data augmentation and  label-smoothing~\cite{szegedy2016rethinking}.
Extended versions~\cite{kim2020puzzle, uddin2020saliencymix, kim2021comixup} of CutMix have been designed for realistic mixing using saliency or a proper combination between samples.
They have demonstrated that more realistic augmentations have superior performance, which is related to our motivation to generate realistic and diverse data augmentations for videos.
VideoMix~\cite{yun2020videomix} is an extension of CutMix for video recognition, but the spatial version of VideoMix has shown the best performance.
\cite{dwibedi2020counting} has tried temporally varying augmentation that is closely related to our method. However, their target scopes are limited to the specific task, and they lacked a comprehensive analysis of augmentation parameters.

Because the aforementioned advanced data augmentation methods have become essential components of the modern training recipes, including CNNs~\cite{bello2021revisiting, wightman2021resnet, liu2022convnet} and Transformers~\cite{touvron2021training}, the development of strong data augmentations for video recognition is highly in demand as a fundamental building unit.

\section{DynaAugment}

\subsection{Problem Definition}
\label{sec:autoda}

Given a video data $\textbf{V} = [I_{1}, I_{2}, ..., I_{T}]$ that contains a total of $T$ frames, applying a single augmentation operation (\texttt{op}) at frame $t$ using an augmentation strength $M$ is $\hat{I}_{t} = \texttt{op}(I_{t}, M)$, that describes a static augmentation whose magnitude is identical over whole video frames.
In DynaAugment, we define how to determine the magnitude array $\textbf{M}=[M_{1}, M_{2}, ..., M_{T}]$ as a function of time $t$ to apply $\hat{I}_{t} = \texttt{op}(I_{t}, M_{t})$ as a main problem beyond the naive extension of static augmentations.

\subsection{DynaAugment}
\label{sec_dynaaug}

\begin{figure*}[!t]
\begin{center}
\resizebox{\linewidth}{!}{
  \begin{tabular}{ccccl} 
         & \rotatebox[origin=c]{45}{Dynamic?} & \rotatebox[origin=c]{45}{Smooth?} & \rotatebox[origin=c]{45}{Diverse?} & \quad \large{Graph of $\textbf{M}$} \quad \quad \quad \quad  \large{Video Frames $\Longrightarrow$} \\
      \hline
      \large{\makecell{Input}}  &   &  & & \parbox[c]{\linewidth}{
        \includegraphics[width=1.0\linewidth]{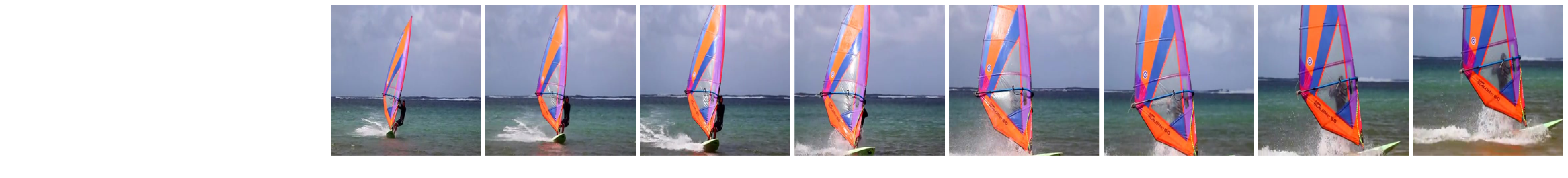}} \\ \cline{1-4}
     \large{\makecell{Static}}  &  \textcolor{red2}{\ding{56}} & \textcolor{darkergreen}{\ding{52}} & \textcolor{red2}{\ding{56}} & \parbox[c]{\linewidth}{
        \includegraphics[width=1.0\linewidth]{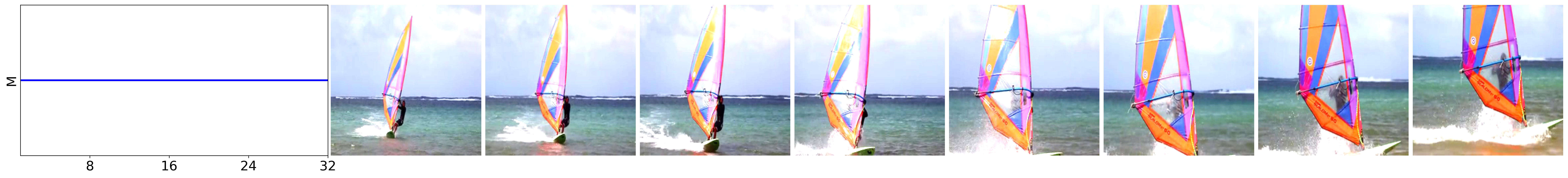}} \\ \cline{1-4}
     \large{\makecell{Random}} & \textcolor{darkergreen}{\ding{52}} & \textcolor{red2}{\ding{56}} & \textcolor{darkergreen}{\ding{52}} & \parbox[c]{\linewidth}{
        \includegraphics[width=1.0\linewidth]{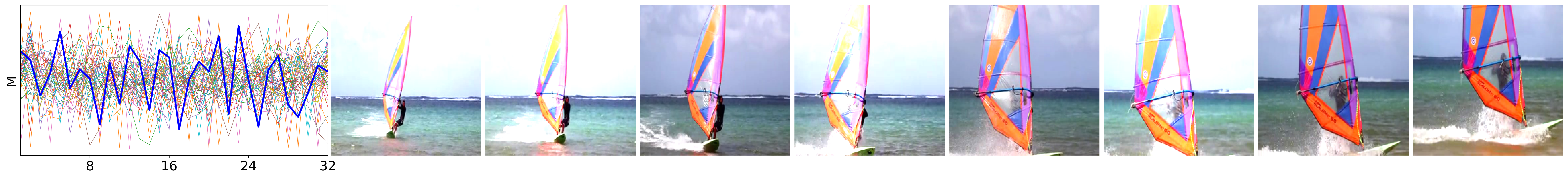}} \\  \cline{1-4}
     \large{DynaAugment} & \textcolor{darkergreen}{\ding{52}} & \textcolor{darkergreen}{\ding{52}} & \textcolor{darkergreen}{\ding{52}} & \parbox[c]{\linewidth}{
        \includegraphics[width=1.0\linewidth]{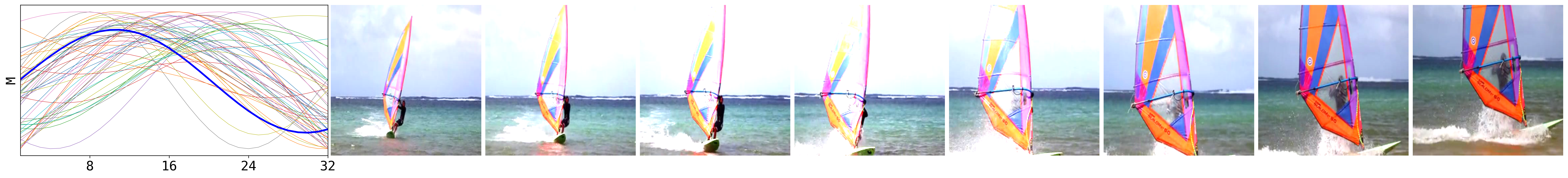}} \\
  \end{tabular}}
  \vspace{-2mm}
    \captionof{figure}{Comparison between static and dynamic augmentations. \textbf{$X$-axis in Graph:} temporal index from the 1st frame to the 32nd frame. \textbf{$Y$-axis in Graph:} magnitude of operations, in this example, \green{\texttt{Brightness}} operation is applied. 
  \textbf{\blue{Blue} Line in Graph:} selected $\textbf{M}$.
  \textbf{Other Lines in Graph:} randomly sampled $\textbf{M}$s in 50 trials.
  \textbf{1st Row:} Frames before augmentation.
  \textbf{2nd Row:} Static Augmentation (same brightness).
  \textbf{3rd Row:} Random (Per-frame) augmentations, it loses temporal consistency.
  \textbf{4th Row:} \textbf{\textit{\mbox{DynaAugment}}} generates dynamic, smooth, and diverse temporal variations (getting \textit{brighter} then \textit{darker} again). Only 8 frames (frame rate is 4) are shown in the figure.
    }\label{fig:mainfig}
  \end{center}
\end{figure*}

In DynaAugment, our motivation to design the dynamic variations is that if a distribution of augmented data is similar to real videos, the generalization performance can be increased.
To implement the realistic augmented videos, we set two constraints in changing the magnitudes.
These constraints and their visualizations are described in Fig.~\ref{fig:mainfig}.
First, an augmentation parameter should be changed as \textit{smoothly} as possible because arbitrary (random) changes can break the temporal consistency.
Second, a set of possible arrays $\textbf{M}$ generated during training should be as \textit{diverse} as possible to increase the representation and generalization space of augmented samples.
The static version (scalar $M$) cannot satisfy both constraints (2nd Row in Fig.~\ref{fig:mainfig}).
Meanwhile, the random variant can generate diverse variations but fails to maintain temporal consistency (3rd Row in Fig.~\ref{fig:mainfig}).
In contrast, DynaAugment (4th Row in Fig.~\ref{fig:mainfig}) is designed for general dynamic data augmentation methodology for videos, as smooth and diverse as possible.
From the perspective of the measurements to quantitatively evaluate augmentation methods proposed in~\cite{gontijo2020tradeoffs}, \textit{smoothness} and \textit{diversity} of ours can be interpreted as the affinity and the diversity of~\cite{gontijo2020tradeoffs}, respectively.
See Sec.~\ref{analysis} for the analysis.

To generate a smooth and diverse temporal array $\textbf{M}$, we propose a novel sampling function called Fourier Sampling, as explained in the next section.
In summary, DynaAugment is an extension of any image augmentation operations (\textit{e.g.} \cite{cubuk2020randaug,muller2021trivialaugment,lingchen2020uniformaugment} in this paper) with the Fourier Sampling to generate generalized temporal arrays ($\textbf{M}$s).

\begin{figure}[!ht]
	\centering
	\begin{subfigure}[c]{0.19\textwidth}
		\centering
        \includegraphics[width=\linewidth]{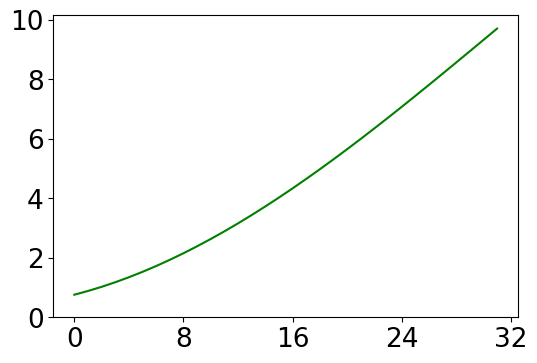}
        \caption{\small{[\blue{0.1},\green{0.2},\sd{0.9},0]}}
	\end{subfigure}
	\hfill
	\begin{subfigure}[c]{0.19\textwidth}
		\centering
        \includegraphics[width=\linewidth]{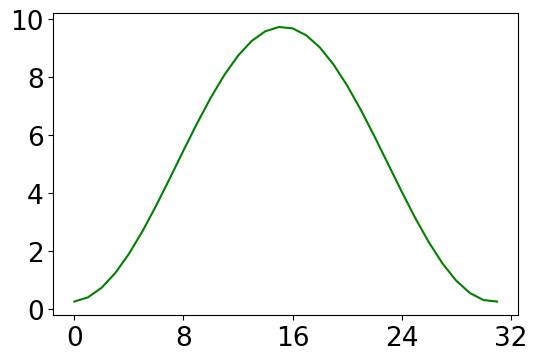}
        \caption{\small{[\blue{0.3},\green{1.0},\sd{0.9},8]}}
	\end{subfigure}
		\hfill
	\begin{subfigure}[c]{0.19\textwidth}
		\centering
        \includegraphics[width=\linewidth]{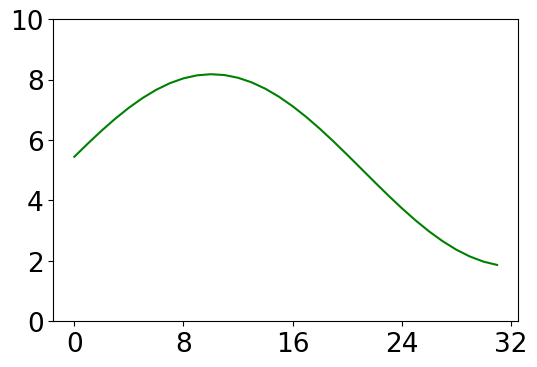}
        \caption{\small{[\blue{0.2},\green{0.7},\sd{0.6},1]}}
	\end{subfigure}
		\hfill
	\begin{subfigure}[c]{0.19\textwidth}
		\centering
        \includegraphics[width=\linewidth]{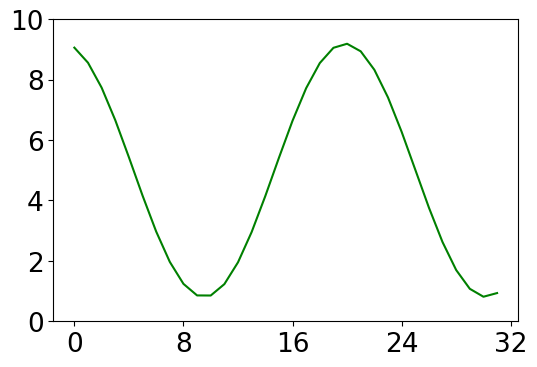}
        \caption{\small{[\blue{0.4},\green{1.5},\sd{0.8},6]}}
	\end{subfigure}
		\hfill
	\begin{subfigure}[c]{0.19\textwidth}
		\centering
        \includegraphics[width=\linewidth]{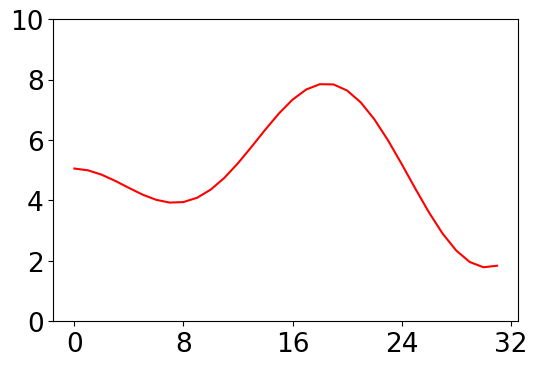}
        \caption{\small{Combined $\textbf{M}$}}
	\end{subfigure}
	\\
\caption{
An example of Fourier Sampling. [\blue{$w_{b}$}, \green{$f_{b}$}, \sd{$A$}, $o_{b}$] of each basis is described.}
	\label{fig:fs}
\end{figure}

\subsection{Fourier Sampling}
\label{sec_fourier}
As mentioned in the introduction, inspired by the Fourier analysis, we make an array $\textbf{M}$ as a weighted sum of $C$ sinusoidal basis functions to generate arbitrary smooth and partially periodic arrays.
\begin{equation}
	\textbf{M} = \sum^C_b w_{b}(\texttt{norm}(\text{sin}[2 f_{b} \pi \textbf{K}[o_{b}:o_{b}+T] / (T-1)]))
    \label{equation:fourier}
\end{equation}
\noindent where $\textbf{K} = [1, \ldots, 2T]$, $T$ is the number of frames, and $[:]$ is an slicing function.
For each basis $b$, $w_{b}$ is a weight sampled from $\texttt{Dirichlet}(1.0)$, 
$f_{b}$ is a frequency that is randomly sampled from $\texttt{Uniform}(0.2, 1.5)$, 
and \texttt{norm} is a normalization operation that is determined by an amplitude $A$, which is sampled from $\texttt{Uniform}(0, 1)$.
From $A$, a signal is min-max normalized within $[M - M * (A / 1.0), M + M * (A / 1.0)]$, where $M$ is a static magnitude.
$o_{b}$ is an offset that shifts the array in the $x$-axis, which is sampled from $[0, \ldots, T-1]$. 
There are many possible combinations of $C$, $f_{b}$, $A$, and $o_{b}$.
Through Eq.~\ref{equation:fourier}, we can obtain a combined array $\textbf{M}$ as an arbitrary and smooth temporal variations.
Fig.~\ref{fig:fs} shows an example of this process.
A combination of diverse basis signals (Fig.~\ref{fig:fs} (a)$\sim$(d)) produces a combined signal (Fig.~\ref{fig:fs} (e)). 
More details behind this setting are in Appendix A.

\section{Experiments} \label{sec:experiments}

\paragraph{Baselines from static data augmentations}

We set automatic augmentation methods used in image data as our baseline for DynaAugment.
This line of work shares the following settings: 
(1) a pool (or search space) of basic augmentation operations (\textit{e.g.} \texttt{PIL} operations used in~\cite{cubuk2019autoaugment}), 
(2) during training, the data-loader selects $N$ operations from the search space with a magnitude $M$ and a probability \textit{$p$} to apply that operation, and 
(3) algorithm-specific way to find the optimal settings of (1) and (2). 

AutoAugment (AA)~\cite{cubuk2019autoaugment} finds optimal sub-policies via reinforcement learning and each sub-policy contains $N=2$ consecutive operations with \textit{$p$} and $M$ for each operation.
Since AA requires tremendous computations even in the image data, searching from scratch for video is a more challenging task to execute.
We exclude these searching-based methods such as AA, Adversarial AA~\cite{zhang2019adversarial}, and Fast AA~\cite{lim2019fast} for implementation simplicity.
Instead, we use more simple methods to implement such as 
RandAugment (RA)~\cite{cubuk2020randaug}, TrivialAugment (TA)~\cite{muller2021trivialaugment}, and UniformAugment (UA)~\cite{lingchen2020uniformaugment} to check the effects of DynaAugment as proof-of-concept.
Specifically, RA~\cite{cubuk2020randaug} uses a grid search on $N$ and $M$ per dataset and model.
In this work, we use searched parameters ($N=2$, $M=9$) from ImageNet as a baseline. See Appendix B for more searched parameters for RA. 
TA~\cite{muller2021trivialaugment} begins from $N=1$ and $p=1.0$, then randomly sample an operations as well as its magnitude $M$. 
Similarly, UA~\cite{lingchen2020uniformaugment} proposes tuning-free settings; instead, it randomly samples operation, \textit{$p$} and $M$ from the uniform distribution.

As discussed in~\cite{lingchen2020uniformaugment, muller2021trivialaugment}, the search space for augmentation operations used in~\cite{cubuk2019autoaugment} may not be optimal. 
They empirically verify that the wider (stronger) search space improves performance.
Similar insight for video data is naturally required.
Therefore, we extend search space with three additional operations: dynamic scale, dynamic color, and dynamic random erase~\cite{zhong2020random}, which are only realized in video data. 
For fair comparison, we also use the extended search spaces for RA, TA, and UA. In this case, the additional operations results in the following operations: random resized crop, color jittering, and static random erasing, respectively. More details are described in Appendix A.

\paragraph{Implementation Details}

We reproduce the models that are trained with diverse data augmentation methods: RA~\cite{cubuk2020randaug}, UA~\cite{lingchen2020uniformaugment}, TA~\cite{muller2021trivialaugment}, and our DynaAugment (DA).
\textit{Note that all other methods except DA are static versions.} 
For RA, TA, and UA, they are identical to DA with zero amplitude.
In Fourier Sampling, we set $C=3$ as default.
We experiment on diverse types of video models including: 2D-CNN~\cite{lin2019tsm, wang2021tdn}, 3D-CNN~\cite{feichtenhofer2019slowfast}, 2+1D-CNN~\cite{xie2018rethinking}, efficient CNN~\cite{feichtenhofer2020x3d}, transformers~\cite{fan2021multiscale, liu2021video}, and transformers with convolutions~\cite{li2022uniformer}.
We compare all methods in: 
\begin{enumerate*}[label={\arabic*)}]
	\item large-scale dataset: Kinetics-400~\cite{carreira2017quo} and Something-Something-v2~\cite{goyal2017something},
	\item small-scale dataset: UCF-101~\cite{soomro2012ucf101} and HMDB-51~\cite{kuehne2011hmdb},
	\item fine-grained dataset: Diving-48~\cite{li2018resound} and FineGym~\cite{shao2020finegym},
	\item and transfer learning tasks: video action localization on THUMOS'14~\cite{THUMOS14}, action segmentation on Breakfast~\cite{Kuehne12breakfast}, and object detection on MOT17det~\cite{milan2016mot16}.
\end{enumerate*}
We set all other settings and hyper-parameters following the default values of each augmentation method and model.
More details are described in Appendix A.
More results for transfer learning are describe in Appendix B.
Due to the space limit, we only describe one augmentation baseline for each downstream task.

\subsection{Results on Kinetics-400 and Something-Something-v2}

\paragraph{Datasets}
Kinetics~\cite{carreira2017quo} is a large-scale dataset used as a representative benchmark for many video action recognition studies.
It is mainly used as a pretrain dataset for many downstream tasks.
Kinetics-400 consists of 240K training and 20K validation videos in 400 action classes.
Something-Something-v2 dataset~\cite{goyal2017something} contains more fine-grained temporal actions.
The dataset contains 169K training and 25K validation videos with 174 action classes.

\begin{table}
    \caption{Results on Kinetics-400 and Something-Something-v2 dataset (Top-1 / Top-5 Accuracy). Each baseline model \textit{does not contain strong augmentations}. ($\ast$: ImageNet pretrained,  $\dagger$: trained from scratch, $\diamond$: Kinetics-400 pretrained, \textbf{RA}: RandAugment, \textbf{TA}: TrivialAugment, \textbf{UA}: UniformAugment, \textbf{DA}: DynaAugment.)}
    \label{tab:main1}
    \centering
    \small
    \resizebox{1.0\textwidth}{!}{
        \begin{tabular}{lcccccccc}
            \toprule
             & \multicolumn{7}{c}{\textbf{Kinetics-400}} \\
            \cmidrule(lr){2-8}
            \textbf{Video Model} & \textbf{Baseline}&\textbf{RA}&\textbf{RA+DA}&\textbf{TA}&\textbf{TA+DA}&\textbf{UA}&\textbf{UA+DA}\\
            \midrule
            TSM-R50-8$\times$8$\ast$~\cite{lin2019tsm} & 74.2 / 91.4 & 74.4 / 91.5 & \textbf{75.1 / 92.2} & 75.5 / 92.4 & \textbf{75.8 / 92.6} & 75.4 / 92.2 & \textbf{75.7 / 92.8} \\
            TDN-R50-8$\times$5$\ast$~\cite{wang2021tdn} & 76.9 / 92.9 & 77.8 / 93.6 & \textbf{77.9 / 93.7} & 78.3 / 93.6 & \textbf{78.5 / 94.0} & 78.2 / 93.7 & \textbf{78.7 / 94.3} \\
             SlowOnly-R50-8$\times$8$\dagger$~\cite{feichtenhofer2019slowfast} & 74.8 / 91.6 & 74.7 / 91.3 & \textbf{75.4 / 92.0} & 75.6 / 92.2 & \textbf{76.1 / 92.7} & 75.7 / 92.1 & \textbf{76.2 / 92.6} \\
             SlowFast-R50-8$\times$8$\dagger$~\cite{feichtenhofer2019slowfast} & 76.8 / 92.6 & 77.3 / 93.2 & \textbf{78.0 / 93.9} & 78.5 / 93.6 & \textbf{79.1 / 94.1} & 78.3 / 93.5 & \textbf{79.0 / 94.0} \\
             X3D-M-16$\times$5$\dagger$~\cite{feichtenhofer2020x3d} & 76.0 / 92.4 & 76.3 / 92.6 & \textbf{76.9 / 93.1} & 76.6 / 92.9 & \textbf{77.0 / 93.3} & 76.6 / 93.1 & \textbf{77.1 / 93.4} \\
             Swin-Tiny-32$\times$2$\ast$~\cite{liu2021video} & 78.9 / 93.9 & 79.4 / 94.2 & \textbf{80.0 / 94.3} & 79.6 / 94.2 & \textbf{80.2 / 94.5} & 79.6 / 94.2 & \textbf{80.1 / 94.4} \\
            \midrule

             & \multicolumn{7}{c}{\textbf{Something-Something-v2}} \\
             \cmidrule(lr){2-8}
             & \textbf{Baseline} &\textbf{RA}&\textbf{RA+DA}&\textbf{TA}&\textbf{TA+DA}&\textbf{UA}&\textbf{UA+DA} \\
            \midrule
             TSM-R50-16$\ast$~\cite{lin2019tsm} & 63.2 / 88.0 & 65.4 / 89.6 & \textbf{66.0 / 89.7} & 65.7 / 89.3 & \textbf{66.1 / 89.8} & 65.7 / 89.5 & \textbf{66.3 / 90.0} \\
             TDN-R50-8$\times$5$\ast$~\cite{wang2021tdn} & 64.0 / 88.8 & 64.9 / 88.4 & \textbf{65.6 / 89.6} & 66.1 / 89.9 & \textbf{66.6 / 90.2} & 66.3 / 89.9 & \textbf{66.6 / 90.2} \\
             SlowFast-R50-8$\times$8$\diamond$~\cite{feichtenhofer2019slowfast} & 61.4 / 85.8 & 63.1 / 87.3 & \textbf{65.0 / 89.0} & 63.5 / 89.0 & \textbf{64.8 / 88.9} & 63.1 / 88.7 & \textbf{64.0 / 88.9} \\
             SlowFast-R50-16$\times$8$\diamond$~\cite{feichtenhofer2019slowfast} & 63.0 / 88.5 & 64.4 / 88.7 & \textbf{65.5 / 89.5} & 64.7 / 88.6 & \textbf{65.8 / 89.8} & 64.6 / 88.5 & \textbf{65.7 / 89.6} \\
            \bottomrule
        \end{tabular}}
\end{table}

\paragraph{Results on models without strong augmentations as default}
Early studies of video backbone models for recognition have generally not used strong data augmentation (\textit{e.g.} RA) or regularization (\textit{e.g.} MixUp) as default recipes.
They only contain standard data augmentations such as random temporal sampling, scale jittering, random crop, and random flip. 
% Random flip is not used in something-something-v2 due to direction-sensitive classes.
We describe experimental results on those models with augmentation methods in Table~\ref{tab:main1}.

In Kinetics-400, using static automatic augmentation consistently improves the performances, for example, TA boosts the performance from 0.6\% (X3D-M) to 1.7\% (SlowFast).
Using dynamic extension, the performances are further boosted up to 0.6\%. 
DA improves the performance by 2.3\% in SlowFast-R50-8x8 that is even  better than the larger model, such as  SlowFast-R101-16×8~\cite{feichtenhofer2019slowfast}.
The improvements are similar in Something-Something-v2, but the improvement gaps became larger. This demonstrates the effect of DA in more dynamic dataset.
Interestingly, the breadth of performance improvement tends to increase as the model's capacity increases, the number of input frames increases, or the model is not pretrained.

\begin{table}[!t]
        \caption{Results on Kinetics-400 dataset (Top-1 / Top-5 Accuracy). Each baseline model \textit{contains strong augmentations as default with RandAugment.} ($\ast$: ImageNet pretrained,  $\dagger$: trained from scratch, $\diamond$: Kinetics-400 pretrained, \textbf{RA}: RandAugment,  \textbf{DA}: DynaAugment.)}
        \label{tab:main2}
        \centering
        \small
        \resizebox{0.85\textwidth}{!}{
            \begin{tabular}{lcccccc}
                \toprule
                & \multicolumn{5}{c}{\textbf{Kinetics-400}} \\
                \cmidrule(lr){2-6}
                \textbf{Video Model} & \textbf{Baseline}&\textbf{Baseline with DA}&\textbf{No Aug.}&\textbf{RA Only}&\textbf{RA+DA Only}\\
                \midrule
                 Swin-Tiny-32$\times$2$\dagger$~\cite{liu2021video} & 77.1 / 93.1 & \textbf{77.6 / 93.3} & 73.9 / 90.5 & 76.4 / 92.6 & \textbf{76.9 / 92.9} \\
                 MViT-S-16$\times$4$\dagger$~\cite{fan2021multiscale} & 76.1 / 92.3 & \textbf{76.7 / 92.9} & 69.4 / 88.0 & 73.0 / 91.0 & \textbf{73.9 / 91.5} \\
                 Uniformer-S-16$\times$8$\ast$~\cite{li2022uniformer} & 80.1 / 94.4 & \textbf{80.9 / 94.8} & 76.6 / 92.2 & 79.5 / 93.9 & \textbf{80.3 / 94.4} \\
                 \midrule
                 & \multicolumn{5}{c}{\textbf{Something-Something-v2}} \\
                \cmidrule(lr){2-6}
                 & \textbf{Baseline}&\textbf{Baseline with DA} & & & \\
                 \midrule
                 Swin-Base-32$\times$2$\diamond$~\cite{liu2021video} & 69.9 / 92.7 & \textbf{70.7 / 93.0} & - & - & - \\
                 Uniformer-B-16$\times$4$\diamond$~\cite{li2022uniformer} & 70.2 / 92.5 & \textbf{71.0 / 93.2} & - & - & - \\
                \bottomrule
            \end{tabular}
            }
    \end{table}
    
\paragraph{Results on models with strong augmentations as default}

Recent models~\cite{liu2021video, fan2021multiscale, li2022uniformer} use a combination of strong augmentations and regularizations. For example, label smoothing~\cite{szegedy2016rethinking}, stochastic depth~\cite{huang2016deep}, random erasing~\cite{zhong2020random}, MixUp~\cite{zhang2017mixup}, CutMix~\cite{yun2019cutmix}, RandAugment~\cite{cubuk2020randaug}, and repeated augmentation~\cite{hoffer2020augment} are used.
From this recipe, we extend RandAugment into DynaAugment but maintain other augmentations.
Note that Swin-Tiny in Table~\ref{tab:main2} is trained from scratch different from the original setting.

The results in Table~\ref{tab:main2} show that the clear improvements by dynamic extensions in all cases from 0.5\% to 0.8\% Top-1 accuracy. 
We also conduct experiments without other augmentations but only RA is used (the right side of Table~\ref{tab:main2}).
From the results, RA greatly improves the performance from the \textit{no-aug} baseline, and DA further boosts the performances from 0.5\% to 0.9\%.

\subsection{Results on Other Datasets}
\label{sec:ucf}

\begin{table}
        \caption{Results on UCF-101, HMDB-51, Diving-48, and Gym288 dataset (Top-1 / Top-5 Accuracy). $^{\ddag}$: Per-class evaluation. (\textbf{RA}: RandAugment, \textbf{UA}: UniformAugment, \textbf{TA}: TrivialAugment, \textbf{DA}: DynaAugment.)}
        \label{tab:small}
        \centering
        \small
        \resizebox{0.95\textwidth}{!}{
            \begin{tabular}{lcccccccc}
                \toprule
                \textbf{Dataset} & \textbf{Baseline}&\textbf{RA}&\textbf{RA+DA}&\textbf{TA}&\textbf{TA+DA}&\textbf{UA}&\textbf{UA+DA}\\
                \midrule
                UCF-101~\cite{soomro2012ucf101} & 72.0 / 90.6 & 82.8 / 96.4 & \textbf{85.3 / 97.2} & 81.6 / 96.3 & \textbf{82.8 / 95.7} & 82.5 / 96.1 & \textbf{82.9 / 95.6} \\
                HMDB-51~\cite{kuehne2011hmdb} & 40.1 / 72.6 & 48.5 / 78.4 & \textbf{54.8 / 83.7} & 47.7 / 78.9 & \textbf{51.9 / 80.7} & 49.9 / \textbf{79.8} & \textbf{52.4} / 78.9 \\
                Diving-48~\cite{li2018resound} & 62.8 / 93.6 & 63.6 / 93.8 & \textbf{70.5 / 95.6} & 68.3 / 95.1 & \textbf{72.5 / 95.7} & 67.8 / 95.0 & \textbf{70.0 / 94.9} \\
                Gym288$^{\ddag}$~\cite{shao2020finegym} & 54.9 / 83.9 & 52.6 / 80.6 & \textbf{55.2 / 82.0} & 54.4 / 80.9 & \textbf{55.6 / 83.2} & 53.4 / 82.4 & \textbf{56.1 / 83.5} \\
                \bottomrule
            \end{tabular}
            }
\end{table}
    
\begin{figure}[!t]
	\centering
	\begin{subfigure}[c]{0.4\textwidth}
		\centering
		\includegraphics[width=\linewidth]{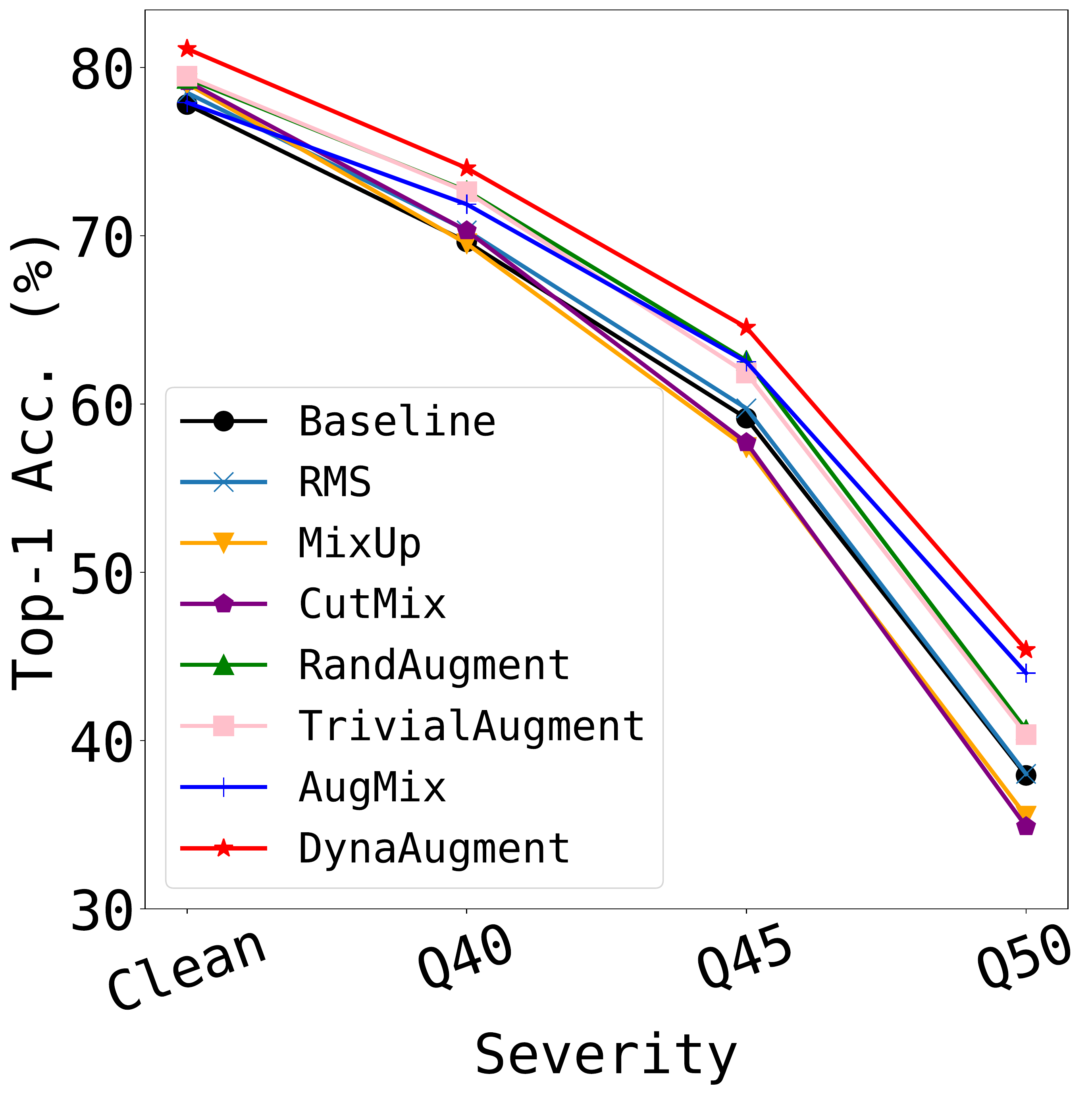}
		\caption{Severity-Performance Plot}
	\end{subfigure}
	\hfill
	\begin{subfigure}[c]{0.55\linewidth}
    	\centering
    	\includegraphics[width=\linewidth]{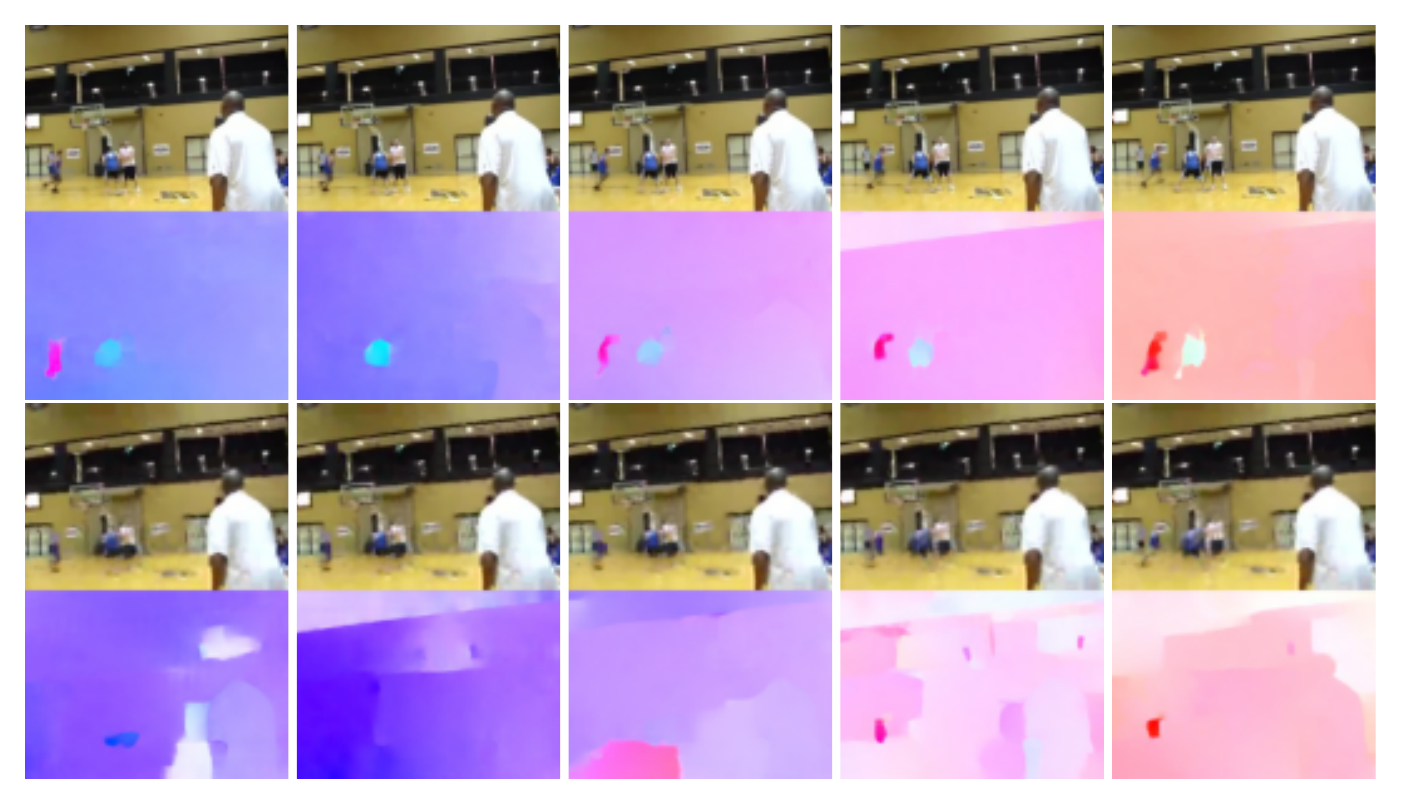}
    	\caption{Visualization}
	\end{subfigure}
	\\
\caption{
(a) Performance drop on spatio-temporally corrupted videos. \texttt{Q} indicates the quantization parameter of compression, and high \texttt{Q} indicates severe degradation. (b) Visualization of temporal variations using optical flow. Five frames above: original video, below: compressed video.
}
\label{fig:corruption_results}
\end{figure}

\paragraph{Datasets and Settings}

We use UCF-101~\cite{soomro2012ucf101} and HMDB-51~\cite{kuehne2011hmdb} datasets for verifying the effectiveness of the data augmentations in the small-scale dataset.
UCF-101~\cite{soomro2012ucf101} contains 13,320 trimmed videos in 101 action classes.
HMDB-51~\cite{kuehne2011hmdb} contains 6,849 videos in 51 action classes.
Both dataset consist of three train/test splits, and we use the first split.

Diving-48~\cite{li2018resound} and FineGym~\cite{shao2020finegym} datasets are used as fine-grained action recognition.
Diving-48~\cite{li2018resound} dataset contains 18,404 videos in 48 fine-grained diving action classes. 
Gym288 dataset in FineGym~\cite{shao2020finegym} contains 38,980 videos in 288 fine-grained gymnastic action classes.
Since all videos share similar background and static contexts, the model should have strong temporal understanding to catch the fine-grained action of an athlete~\cite{choi2019can}.
For Gym288, because the distribution of the dataset is heavily imbalanced, we evaluate per-class accuracy. 
The S3D-G~\cite{xie2018rethinking} model is used and trained from scratch to check the clear effect of data augmentation for all experiments.

\paragraph{Results}

Table~\ref{tab:small} shows the results on the four datasets.
The results show that DA consistently outperforms its static counterpart.
Specifically, in UCF-101 dataset, being dynamic over RA improves up to 2.5\% in top-1 accuracy, and 13.3\% over baseline. The results in HMDB-51 show a similar tendency.
In Diving-48 dataset, DA increase the performance by 6.9\% over RA that indicate DA also significantly improves fine-grained representations of the model.
The results indicate that better data augmentation is critical to the performance improvements in the small-scale dataset.
In Gym288 dataset, the results are not effective as in the other datasets; static augmentation even degrades the baseline performance. Nevertheless, DA improves the accuracy up to 1.2\%.

\subsection{Results on Corrupted Videos.}

To test and compare the corruption robustness of the data augmentation algorithms in video recognition, we make a validation dataset from the subset of Kinetics through \textit{video compression}.
In a mobile or streaming environment, the quality of videos can become extremely degraded due to the network conditions, and it can be dynamic (See Fig.~\ref{fig:corruption_results} (b)) due to the adaptive rate control. 
We use H.264/AVC~\cite{wiegand2003overview} to compress the videos with different levels of QP (quantization parameters).

Fig.~\ref{fig:corruption_results} (a) presents the results as function of each corruption's severity. 
Automated data augmentation such as RA~\cite{cubuk2020randaug} and TA~\cite{muller2021trivialaugment} show good robustness over baseline and regularization methods including RMS~\cite{kim2020regularization}, MixUp~\cite{zhang2017mixup} and CutMix~\cite{yun2019cutmix}.
AugMix~\cite{hendrycks2019augmix} is designed for the corruption robustness and also shows good performances in hard corruptions, but it falls behind with clean and less corrupted inputs.
In contrast, DA shows best results over every severity level demonstrating good clean accuracy and corruption robustness.
Because in the corrupted videos, it is observed that the spatio-temporal variations between frames are generally increased as shown in Fig.~\ref{fig:corruption_results} (b), augmented representation of DA can be generalized into corrupted videos.

\subsection{Transfer Learning}

\begin{table}
        \caption{Transfer learning to video action localization task. Results on THUMOS'14 dataset. (\textbf{UA}: UniformAugment, \textbf{TA}: TrivialAugment,  \textbf{DA}: DynaAugment.)}
        \label{tab:thumos}
        \centering
        \small
        \resizebox{1.0\textwidth}{!}{
            \begin{tabular}{lccccccccc}
                \toprule
                & \multicolumn{8}{c}{\textbf{Fully-Supervised Learning on G-TAD}} \\
                \cmidrule(lr){2-9}
                \textbf{Configuration} & mAP 0.1 &	mAP 0.2 &	mAP 0.3 &	mAP 0.4 &	mAP 0.5 &	mAP 0.6 &	mAP 0.7 &	Avg mAP \\
                \midrule
                SlowOnly-50 & 58.7 & 55.2 & 51.1 & 44.2 & 34.2 & 24.7 & 15.3 & 33.9 \\
                SlowOnly-50+UA & 60.8 & 57.7 & 53.4 & 46.2 & 36.8 & 27.1 & 17.4 & 36.2 \\
                SlowOnly-50+UA+\textbf{DA} & \textbf{61.5} & \textbf{58.1} & \textbf{53.8} & \textbf{47.0} & \textbf{37.8} & \textbf{28.2} & \textbf{18.5} & \textbf{37.1} \\
                
                \midrule
                & \multicolumn{8}{c}{\textbf{Weakly-Supervised Learning on BaSNet}} \\
                \cmidrule(lr){2-9}
                \textbf{Configuration} & mAP 0.1 &	mAP 0.2 &	mAP 0.3 &	mAP 0.4 &	mAP 0.5 &	mAP 0.6 &	mAP 0.7 &	Avg mAP \\
                \midrule
                SlowOnly-50 & 51.7 & 46.0 & 37.9 & 31.1 & 23.9 & 15.6 & 8.3 & 30.6 \\
                SlowOnly-50+TA & 52.3 & 46.9 & 39.7 & 32.1 & 25.1 & 16.9 & 9.5 & 31.8 \\
                SlowOnly-50+TA+\textbf{DA} & \textbf{52.4} & \textbf{47.2} & \textbf{40.5} & \textbf{33.3} & \textbf{25.3} & \textbf{17.5} & \textbf{9.6} & \textbf{32.3} \\
                \bottomrule
            \end{tabular}
            }
\end{table}

\begin{table}[t!]
	\centering
	\caption{Transfer learning to video action segmentation and video object detection tasks. (\textbf{UA}: UniformAugment, \textbf{TA}: TrivialAugment, \textbf{DA}: DynaAugment.)}\label{tab:ablation}
	
\begin{subtable}[t]{.5\textwidth}
\centering
\caption{Action segmentation results on Breakfast dataset.}\label{tab:seg}
\setlength{\tabcolsep}{3pt}
\small
\vspace{-0.5mm}

\small
\begin{tabular}{lccccc}
    \toprule
        &   & \multicolumn{3}{c}{F1}  \\
    \cmidrule{3-5}
    \textbf{Configuration} & Acc. & @0.10 & @0.25 & @0.50 \\
    \midrule
    SlowOnly-50 & 59.0 & 54.7 & 49.2 & 37.6 \\
    SlowOnly-50+UA & 62.3 & 59.1 & 53.7 & 40.8 \\
    SlowOnly-50+UA+\textbf{DA} & \textbf{64.7} & \textbf{60.6} & \textbf{55.5} & \textbf{43.2} \\
    \bottomrule
\end{tabular}

\end{subtable}
\begin{subtable}[t]{.49\textwidth}

\caption{Object detection results on MOT17det dataset.}\label{tab:det}
\begin{center}
\small
\vspace{-2.3mm}
\setlength{\tabcolsep}{4pt}

\small
\begin{tabular}{lccccc}
    \toprule
    \textbf{Configuration} & AP & AP50 & AP75 & APs \\
    \midrule
    Swin-T & {{30.3}\ms{0.9}\hfill} & 63.7 & 25.1 & 4.1 \\
    Swin-T+TA & {{29.1}\ms{1.3}\hfill} & 62.1 & 23.2 & 3.7 \\
    Swin-T+TA+\textbf{DA} & \textbf{{{31.3}\ms{0.6}\hfill}} & \textbf{65.3} & \textbf{26.3} & \textbf{5.7} \\
    \bottomrule
\end{tabular}
\end{center}
\end{subtable}
\end{table}

\paragraph{Video Action Localization}
THUMOS’14 dataset~\cite{THUMOS14} has 20 class subsets of the untrimmed videos.
The videos contain long sequences (up to 26 min) with sparse action segments, and the goal is to classify video-level actions and localize the temporal actions.
We experiment under two settings: (1) fully-supervised setting on G-TAD~\cite{xu2020g}, and (2) weakly-supervised setting on BaSNet~\cite{lee2020background}.
The features are extracted from Kinetics-400 pretrained SlowOnly-50.
The results in Table~\ref{tab:thumos} show that the performance is  improved by the pretrained model using DA training compared to the model using UA training.
This indicates that DA learns better representations in terms of temporal sensitivity from temporal variations.

\paragraph{Video Action Segmentation}
Breakfast dataset~\cite{Kuehne12breakfast} contains 1,712 untrimmed videos with temporal annotations of 48 actions related to breakfast preparation.
Unlike action localization, all frames are densely classified into pre-defined class. % including background?
We experiment using MS-TCN~\cite{farha2019ms} from the features that are extracted from Kinetics-400 pretrained SlowOnly-50.
Similar to action localization, the results in Table~\ref{tab:seg} demonstrate the superior performance of DA-pretrained features.

\paragraph{Video Object Detection}

For video object detection, MOT17 detection benchmark~\cite{milan2016mot16} is used.
We follow the previous practice to split the MOT17 training set into two parts, one for training and the other for validation.
As a downstream task of the video recognition, from the CenterNet~\cite{zhou2019objects}-based detection framework, we substitute the backbone to our 3D video backbone.
Swin-Tiny is used as our 3D backbone, and the results in Table~\ref{tab:det} show that the detection performance (AP) is clearly increased, especially in the hard cases (AP75 and APs).
With repeated experiments, baseline (Swin-Tiny) and TA show results with high variability confirmed by the high standard deviation (std) of results.
We attribute the inferiority of TA to this variability, with full results available in Appendix B.
Static augmentation may not be proper for some datasets, as observed in Gym288 result of Table~\ref{tab:small}.
Nevertheless, DA not only gives consistently superior results, but the std is also reduced compared to baseline and TA, also demonstrating DA's robustness across datasets.

\subsection{Ablation Study and Discussion}
\label{analysis}

% \paragraph{Search Space}
\noindent{\textbf{Search Space}} As mentioned above, augmentation spaces used for image recognition may not be optimal for video due to the video's dynamic nature.
Table~\ref{tbl:abl} (a) shows the result of different augmentation search spaces in UCF-101 and Kinetics-100 (used in~\cite{chen2021rspnet}) datasets. 
\textit{Org.} means the search space used in the original algorithm and \textit{Mod.} means the extended search space for DA.
\textit{Wide} means the wide search space used in TA. 
The results show that the improvements of DA are not due to the modified search space; instead, making temporally dynamic is the more critical factor.

% \paragraph{Smoothness}
\noindent{\textbf{Smoothness}} As shown in Fig.~\ref{fig:mainfig}, smoothness and diversity are important factors in designing the natural temporal variations. 
In Table~\ref{tbl:abl} (b), the result shows that although static augmentation boosts the performance in UCF-101 and Diving-48, dynamic variations further boost the performance. 
However, simple dynamic extensions such as naive linear or sinusoidal lack diversity, and random variation lacks smoothness. 
These examples are visualized in Appendix A.
In UCF-101, they underperform the static version.
In contrast, in the more temporally-heavy Diving-48 dataset, making variations dynamic is more critical.

% \paragraph{Affinity and Diversity}
\noindent{\textbf{Affinity and Diversity}} To check why the performances of DynaAugment are remarkable, two qualitative measures for data augmentation proposed in~\cite{gontijo2020tradeoffs} are used. 
The first one is the \textit{affinity} which describes the distribution shift from the clean dataset, and the second one is the \textit{diversity} which describes the number of unique training samples. See Appendix A for more details.
The results in Table~\ref{tbl:abl} (c) show that DA has both high affinity and diversity. Meanwhile, the static version has low affinity, the naive linear version has low diversity, and the random version has very low affinity. 
This indicates that the dynamic variations produced by Fourier Sampling are quite natural.

\begin{table}[t]
\caption{Ablation studies and analysis on diverse augmentation settings. For (b) and (c), RA is used. TA is used for Diving-48 in (b).}
\label{tbl:abl}
\begin{subtable}{0.58\linewidth}
\centering
\small
\caption{Results on different search space. Org.: original image search space, Mod.: modified video search space, Wide: wide search space used in TA.}
\begin{tabular}{lllc}
    \toprule
    Dataset & Config. & Space & Top-1/Top-5 \\
    \midrule
    UCF-101 & Baseline & - & 72.0 / 92.6 \\
      & RA & Org. & 79.7 / 94.3 \\
    & RA+DA & Org. & \textbf{83.0 / 96.2} \\
    \cmidrule{2-4}
      & RA & Mod. & 82.8 / 96.4 \\
      & RA+DA & Mod. & \textbf{85.3 / 97.2} \\
     \midrule
      Kinetics-100 & Baseline & - & 66.6 / 85.5 \\
      & TA & Org. & 71.1 / 88.4 \\
      & TA & Wide & 72.3 / 89.4 \\
    & TA+DA & Wide & \textbf{73.2 / 90.0} \\
    \cmidrule{2-4}
      & TA & Wide+Mod. & 72.7 / 89.9 \\
      & TA+DA & Wide+Mod. & \textbf{73.7 / 90.5} \\
    \bottomrule
\end{tabular}
\end{subtable}
\begin{subtable}{0.42\linewidth}
\centering
\small
\caption{Results on different dynamic variations.}
\begin{tabular}{llc}
    \toprule
       Config. & UCF-101 & Diving-48 \\
       \midrule
       Baseline & 72.0 / 92.6 & 62.8 / 93.6 \\
       Static & 82.8 / 96.4 & 68.3 / 95.1 \\
       Linear & 80.1 / 94.8 & 69.2 / \textbf{95.7} \\
       Sinusoidal & 81.4 / 95.5 & 71.8 / 94.9 \\
       Random & 76.9 / 91.7 & 71.0 / 94.0 \\
       Ours & \textbf{85.3 / 97.2} & \textbf{72.5 / 95.7} \\
    \bottomrule
\end{tabular}

\vspace{0.5cm}

\caption{Affinity (Aff.) and diversity (Div.) measurement on diverse dynamic variations. Measured on UCF-101.}
\begin{tabular}{lllc}
    \toprule
    Config. & Aff. & Div. & $\Delta$Top-1. \\
    \midrule
    Static & 0.93 & 1.43 & +10.8 \\ % Low A, High D
    Linear & \textbf{0.96} & 1.39 & +7.9 \\ % High A, Low D
    Random & 0.63 & 1.46 & +4.9 \\ % Very Low A, High D
    Ours & \textbf{0.96} & \textbf{1.59} & \textbf{+13.3} \\ % High A, High D
    \bottomrule
\end{tabular}
\end{subtable}%
\end{table}
\section{Conclusion and Limitation} \label{sec:conclusion}

Extensive experimental results indicate that the key factor in increasing video recognition performance is to make the augmentation process as temporally dynamic, smooth, and diverse as possible.
We implement such properties as a novel framework called DynaAugment. It shows that there is room for performance to be improved on various video recognition tasks, models, and datasets over the static augmentations. 
We expect DynaAugment to be an important component of improved training recipes for video recognition tasks.
Lack of finding these recipes combined with other augmentation and regularization methods is one of our limitation in this paper. 
We reserve it for future work because it requires considerable effort and resources.

{\small
\bibliographystyle{IEEEtran}
\bibliography{IEEEabrv.bib,main}
}

\newpage
\appendix

\setcounter{table}{0}
\renewcommand{\thetable}{A\arabic{table}}
\renewcommand{\thefigure}{A\arabic{figure}}

\section{Implementation Details}

\subsection{Detailed Configurations}

\begin{table}[h]
    \caption{Detailed Configurations for Kinetics-400 models. SGD(M) is SGD with momentum optimizer, the numbers in step schedule are epochs for decaying. For train/test sampling, dense sampling uniformly samples in the whole temporal range, while segment sampling extracts frames from each local segment. For test view, spatial$\times$temporal cropping is used. For the official and reproduced results, Top-1 accuracies (\%) are described.}
    \label{tab:detail_k400}
    \centering
    \small
    \resizebox{1.0\textwidth}{!}{
        \begin{tabular}{l|cccccc|ccc}
            \toprule
            \textbf{Video Model} & TSM~\cite{lin2019tsm}&TDN~\cite{wang2021tdn}&SlowOnly~\cite{feichtenhofer2019slowfast}&SlowFast~\cite{feichtenhofer2019slowfast}&X3D-M~\cite{feichtenhofer2020x3d}&Swin-T~\cite{liu2021video}&Swin-T~\cite{liu2021video}&MViT-S~\cite{fan2021multiscale}&Uniformer-S~\cite{li2022uniformer}\\
            \midrule
            Pretrain & ImageNet & ImageNet & - & - & - & ImageNet & - & - & ImageNet \\
            Frame/Rate & 8$\times$8	& 40	& 8$\times$8	& 32$\times$2	& 16$\times$5	& 32$\times$2	& 32$\times$2 &	16$\times$4	& 16$\times$8 \\
            Epochs	& 150	& 150	& 300	& 300	& 300	& 60	& 200	& 200	& 110 \\
            Batch Size &	64 &	128 &	64 &	128 &	128 &	64 &	64 &	128 &	32 \\
            Learning Rate &	0.01 &	0.02 &	0.2 &	0.2 & 	0.2 & 	1e-03 &	4e-04 &	8e-04 &	1e-04 \\
            Optimizer & SGD(M) &	SGD(M) &	SGD(M) &	SGD(M) &	SGD(M) & AdamW~\cite{loshchilov2017decoupled} &	AdamW~\cite{loshchilov2017decoupled} &	AdamW~\cite{loshchilov2017decoupled} &	AdamW~\cite{loshchilov2017decoupled} \\
            Schedule & Step [50, 100] &	Step [50, 100 ,125] & Cosine~\cite{loshchilov2016sgdr} & Cosine~\cite{loshchilov2016sgdr} &	Cosine~\cite{loshchilov2016sgdr} & 	Cosine~\cite{loshchilov2016sgdr} & Cosine~\cite{loshchilov2016sgdr} & Cosine~\cite{loshchilov2016sgdr} & Cosine~\cite{loshchilov2016sgdr} \\
            Warmup~\cite{goyal2017accurate} &	-	& -	& 40	& 40	& 35	& 5	& 30	& 30	& 10 \\
            Weight Decay & 1e-04 & 1e-04 & 1e-04 & 1e-04 & 5e-05 & 0.02 & 0.05 & 0.05 & 0.05 \\
            DropPath~\cite{huang2016deep} & - & - & - & - & - & 0.1 & 0.1 & 0.1 & 0.1 \\
            DropOut~\cite{srivastava2014dropout} & 0.5 & 0.5 & 0.5 & 0.5 & 0.5 & - & 0.5 & 0.5 & - \\
            \midrule
            Train Sampling &	Dense &	Segment &	Dense &	Dense &	Dense &	Dense &	Dense &	Dense &	Dense \\
Test Sampling &	Dense &	Segment &	Dense &	Dense &	Dense &	Dense &	Dense &	Dense &	Dense \\
Test View &	1$\times$10 &	3$\times$10 &	3$\times$10 &	3$\times$10 &	3$\times$10 &	3$\times$4 &	3$\times$4	 & 1$\times$5 &	1$\times$4 \\
            \midrule
            Repeat Aug.~\cite{hoffer2020augment} &	- &	- &	- &	- &	- &	- &	2 &	2 &	2 \\
RandAug~\cite{cubuk2020randaug} &	- &	- &	- &	- &	- &	- &	(7, 4) &	(7, 4) &	(7, 4) \\
MixUp~\cite{zhang2017mixup} &	- &	- &	- &	- &	- &	- &	0.8 &	0.8 &	0.8 \\
CutMix~\cite{yun2019cutmix} &	- &	- &	- &	- &	- &	- &	1.0 &	1.0 &	1.0 \\
Label Smoothing~\cite{hoffer2020augment} &	- &	- &	- &	- &	- &	- &	0.1	 & 0.1	 & 0.1 \\
            \midrule
            Official &	74.1 &	76.6 &	74.8 &	77.0 &	76.0 &	78.8 &	N/A &	76.0 &	80.8 \\
            Reproduced	 & 74.2	 & 76.9	 & 74.8  &	76.8  &	76.0  &	78.9 &	77.1 &	76.1 &	80.1 \\
            \bottomrule
        \end{tabular}}
\end{table}

We use PyTorch~\cite{paszke2017automatic} for the experiments.
All our experiments are conducted in a single machine with 8 $\times$ A100 or 8 $\times$ V100 GPUs.

We try to follow the original training recipes except for the training epochs for all models.  
This choice follows the previous observations that strong augmentation requires more training iterations for convergence.
In Table~\ref{tab:detail_k400}, we describe all hyper-parameters and compare our reproduced baselines with the official results in Kinetics-400~\cite{carreira2017quo} dataset.
These slight differences might be caused by some modifications, such as batch size (or batch per GPU) or training epochs. However, we show the generalization effects of DynaAugment that boost all performances across the baselines.

In Table~\ref{tab:detail_others}, we describe all hyper-parameters and compare our reproduced baselines with the official results in the other datasets, such as Something-Something-v2~\cite{goyal2017something}, UCF-101~\cite{soomro2012ucf101}, HMDB-51~\cite{kuehne2011hmdb}, Diving-48~\cite{li2018resound}, and Gym288~\cite{shao2020finegym}. Note that UCF-101 and HMDB-51 shares all settings, and Diving-48 and Gym288 also shares all settings.
For the small dataset, such as UCF-101, HMDB-51, Diving-48, and Gym288, we intentionally train the models from scratch to check the clear effects of the data augmentation.
For improved results on these datasets (UCF/HMDB) using ImageNet or Kinetics pre-training, please refer to Section B.3.

\begin{table}[!h]
    \caption{Detailed Configurations for Something-Something-v2~\cite{goyal2017something} (Sth-v2), UCF-101~\cite{soomro2012ucf101}, HMDB-51~\cite{kuehne2011hmdb}, Diving-48~\cite{li2018resound} (DV), and Gym288~\cite{shao2020finegym} (Gym) models. SGD(M) is SGD with momentum optimizer, the numbers in step schedule are epochs for decaying. For train/test sampling, dense sampling uniformly samples in the whole temporal range, while segment sampling extracts frames from each local segment. For test view, spatial$\times$temporal cropping is used. For the official and reproduced results, Top-1 accuracies (\%) are described.}
    \label{tab:detail_others}
    \centering
    \small
    \resizebox{1.0\textwidth}{!}{
        \begin{tabular}{l|cccccc|cc}
            \toprule
            \textbf{Video Model} & TSM~\cite{lin2019tsm}&TDN~\cite{wang2021tdn}&SlowFast~\cite{feichtenhofer2019slowfast}&SlowFast~\cite{feichtenhofer2019slowfast}&Swin-B~\cite{liu2021video}&Uniformer-B~\cite{li2022uniformer}&S3D-G~\cite{xie2018rethinking}&S3D-G~\cite{xie2018rethinking}\\
            \midrule
            Dataset & Sth-v2 & Sth-v2 &Sth-v2 &Sth-v2 &Sth-v2 &Sth-v2 & UCF/HMDB & DV/Gym \\
            \midrule
            Pretrain & ImageNet &	ImageNet & 	K400 & 	K400 &	K400 & 	K400 & 	- &	-  \\
            Frame/Rate & 16 &	40 &	32$\times$2 &	64$\times$2 &	32$\times$2 &	16$\times$4 &	32$\times$2 &	32$\times$2 \\
            Epochs	& 90 &	80 &	40	 & 40 &	60 &	60 &	200 &	200 \\
            Batch Size &	64 &	128	& 64	& 64 &	32 &	40 &	64 &	64  \\
            Learning Rate &	0.01 &	0.02 &	0.12 &	0.12 & 	3e-04 & 	2.5e-04 &	0.025 &	0.025 \\
            Optimizer & SGD(M) &	SGD(M) &	SGD(M) &	SGD(M) &	AdamW~\cite{loshchilov2017decoupled} & AdamW~\cite{loshchilov2017decoupled} &	SGD(M) &	SGD(M)  \\
            Schedule & Step [30, 60] &	Step [40, 60, 70] & Step [26, 33] & Step [26, 33] &	Cosine~\cite{loshchilov2016sgdr} & 	Cosine~\cite{loshchilov2016sgdr} & Cosine~\cite{loshchilov2016sgdr} & Cosine~\cite{loshchilov2016sgdr} \\
            Warmup~\cite{goyal2017accurate} &	-	& -	& 0.19	& 0.19	& 2.5	& 5	& -	& -	 \\
            Weight Decay & 1e-04 & 5e-04 & 1e-06 & 1e-06 & 0.05 & 0.05 & - & - \\
            DropPath~\cite{huang2016deep} & - & - & - & - & 0.4 & 0.4 & - & - \\
            DropOut~\cite{srivastava2014dropout} & 0.5 & 0.5 & 0.5 & 0.5 & - & - & 0.8 & 0.8 \\
            \midrule
            Train Sampling &	Segment &	Segment &	Segment &	Segment &	Segment &	Segment &	Dense &	Segment \\
Test Sampling &	Segment &	Segment &	Segment &	Segment &	Segment &	Segment &	Dense &	Segment \\
Test View &	3$\times$2 &	3$\times$1 &	3$\times$1 &	3$\times$1 &	3$\times$1 &	3$\times$1 &	3$\times$10	 & 3$\times$1 \\
            \midrule
            Repeat Aug.~\cite{hoffer2020augment} &	- &	- &	- &	- &	- &	2 &	- &	- \\
RandAug~\cite{cubuk2020randaug} &	- &	- &	- &	- &	(7, 4) &	(7, 4) &	- & - \\
MixUp~\cite{zhang2017mixup} &	- &	- &	- &	- &	- &	0.8 &	- & -  \\
CutMix~\cite{yun2019cutmix} &	- &	- &	- &	- &	- &	1.0 &	- &	- \\
Label Smoothing~\cite{hoffer2020augment} &	- &	- &	- &	- &	0.1 &	0.1 &	-	 & - \\
            \midrule
            Official &	63.4 &	64.0 &	N/A &	63.0 &	69.6 &	70.4 &	N/A &	N/A  \\
            Reproduced	 & 63.2 &	64.0 &	61.4 &	63.0 &	69.6 &	70.2 &	-	 & - \\
            \bottomrule
        \end{tabular}}
\end{table}

\subsection{Transfer Learning Details}

For all transfer learning experiments, from the recognition models that are pre-trained with or without DynaAugment (or TrivialAugment/UniformAugment), features are pre-extracted. 
From the features, all localization (segmentation or detection) head parts are trained with the downstream datasets.

For all transfer learning algorithms, such as G-TAD~\cite{xu2020g}, BaSNet~\cite{lee2020background}, and MS-TCN~\cite{farha2019ms}, we follow the original implementation without configuration changes.
One exception is the feature extraction stage we describe below.

For G-TAD~\cite{xu2020g}, features are extracted in every frame via sliding window whose window size is 9 frames.
For BaSNet~\cite{lee2020background}, features are extracted in every non-overlapping 16 frames with 25 frame-per-second.
For MS-TCN~\cite{farha2019ms}, features are extracted in every frame via sliding window whose window size is 21 frames, with 15 frame-per-second.

For video object detection, we use CenterNet~\cite{zhou2019objects} as a detection head. The backbone is replace with our video backbones.
The training losses are identical to~\cite{zhou2019objects}, but the configurations are changed: batch size 8, total 30 epochs, weight decay 0.05, AdamW~\cite{loshchilov2017decoupled} optimizer, learning rate 1e-04, and step decay 0.1 at 20 epochs.

\subsection{Details on the Extended Search Space}

\begin{figure}[!h]
	\centering
	\begin{subfigure}[c]{0.329\textwidth}
		\centering
		\includegraphics[width=\linewidth]{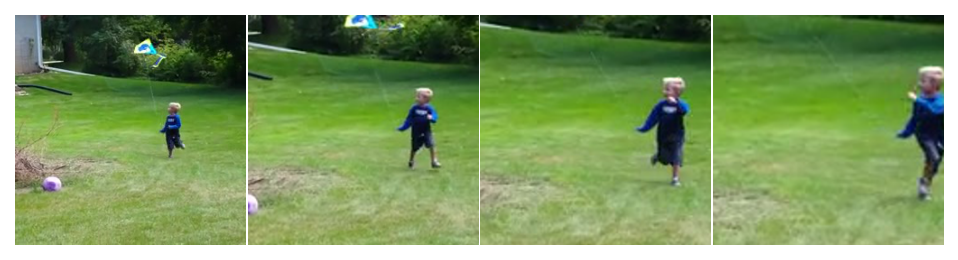}
		\caption{Dynamic Scale}
	\end{subfigure}
	\hfill
	\begin{subfigure}[c]{0.329\linewidth}
    	\centering
    	\includegraphics[width=\linewidth]{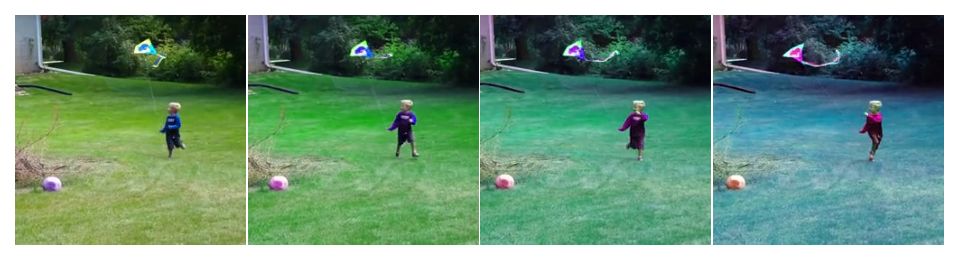}
		\caption{Dynamic Color}
	\end{subfigure}
		\hfill
	\begin{subfigure}[c]{0.329\linewidth}
    	\centering
    	\includegraphics[width=\linewidth]{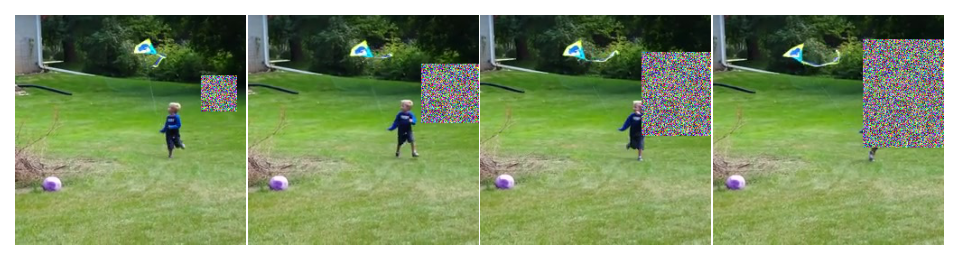}
		\caption{Dynamic Random Erase}
	\end{subfigure}
	\\
\caption{Visualization of augmentation operations used for the extended search space of DynaAugment.
}
	\label{fig:extended}
\end{figure}

As discussed in~\cite{lingchen2020uniformaugment, muller2021trivialaugment}, the search space for augmentation operations used in~\cite{cubuk2019autoaugment} may not be optimal. 
They empirically verify that the wider (stronger) search space improves performance.
Similar insight for video data is naturally required.
Therefore, we extend search space with three additional operations: dynamic scale, dynamic color, and dynamic random erase~\cite{zhong2020random}, which are only realized in video data. 
They are visualized in Fig~\ref{fig:extended}.
For dynamic scale, we first resize an image then add padding (scale-down) or crop centers (scale-up). 
The magnitude range of scale change is [0.667, 1.5], and [0.5, 2.0] for wide search space. 
For dynamic color, we modify an image's hue using \texttt{torchvision} function. The magnitude range is [-0.1, 0.1], and [-0.3, 0.3] for wide.
For dynamic random erase, we follow random erasing in image~\cite{zhong2020random}, but the box size varies across frames.
The magnitude range is [10, 30], and [10, 60] for wide. The numbers in here means the ratio of the area of the box to the total image area.
These three operations are related to the real-world variations, such as, geometric variations (object movements, camera movements, \textit{e.g.}, zoom-in), photometric variations (\texttt{color} in AA~\cite{cubuk2019autoaugment}'s augmentation search space only change the saturation of an image), and time-varying occlusions.

\subsection{Details on Fourier Sampling}

In this subsection, we describe the motivations and procedures of Fourier Sampling introduced in Section. 3.3.
Like a mixture of augmentation described in AugMix~\cite{hendrycks2019augmix}, Fourier Sampling is designed for the mixture of temporal variations.
A basic unit for a temporal variation is a simple periodic signal.
A periodic signal can be varied by modifying its frequencies.

Following~\cite{hendrycks2019augmix}, we first sample weights $w_{b}$ for the basis from the Dirichlet distribution so that the sum of the weight $f_{b}$ becomes 1.
Next, for each basis, we randomly sample a frequency from the uniform distribution between 0.2 to 1.5.
Fig. 3 (a) shows an example signal whose frequency is 0.2.
As shown in the figure, because a signal with $f_{b}=0.2$ is almost linear, the signal is selected as a lower frequency bound for Fourier Sampling. 
Fig. 3 (d) shows an example signal whose frequency is 1.5, and it is selected as a upper bound for Fourier Sampling by the heuristics.
Because most of the video models take up to 32 frames, as described in Section. A.1., too high frequency can hurt temporal consistency.
After sampling a frequency, a sinusoidal signal is generated, and after then, random offset $o_{b}$ and random amplitude $A$ are applied to make the signal more diverse into $x$ and $y$-axis.
An amplitude is proportional to the static magnitude $M$ because more variations are required if a magnitude is large.
Finally, the generated signals are integrated with Eq. (1).
For the number of basis $C$, we experimentally confirm that there was no performance increase for values greater than 3 (See Section B.5.).

% Fig. X. show the visualization of the generated signals as in Fig. 2.
% To check the effects of offset $o_{b}$, see Fig. Y. in Section A.5.
% In Fig. X, if the more basis functions are used, the generated signal becomes diverse.
% Basis C에 따른 갯수 추가.

To achieve the best performance, manual search on the frequency, offset, amplitude, and number of basis is a possible strategy.
However, it costs tremendous computational resources.
Instead of giving it up, we chose a simple random sampling strategy, following search-free methods: TA~\cite{muller2021trivialaugment} or UA~\cite{lingchen2020uniformaugment}.

\subsection{Details on the Smoothness}

\begin{figure}[!ht]
	\centering
	\begin{subfigure}[c]{0.19\textwidth}
		\centering
        \includegraphics[width=\linewidth]{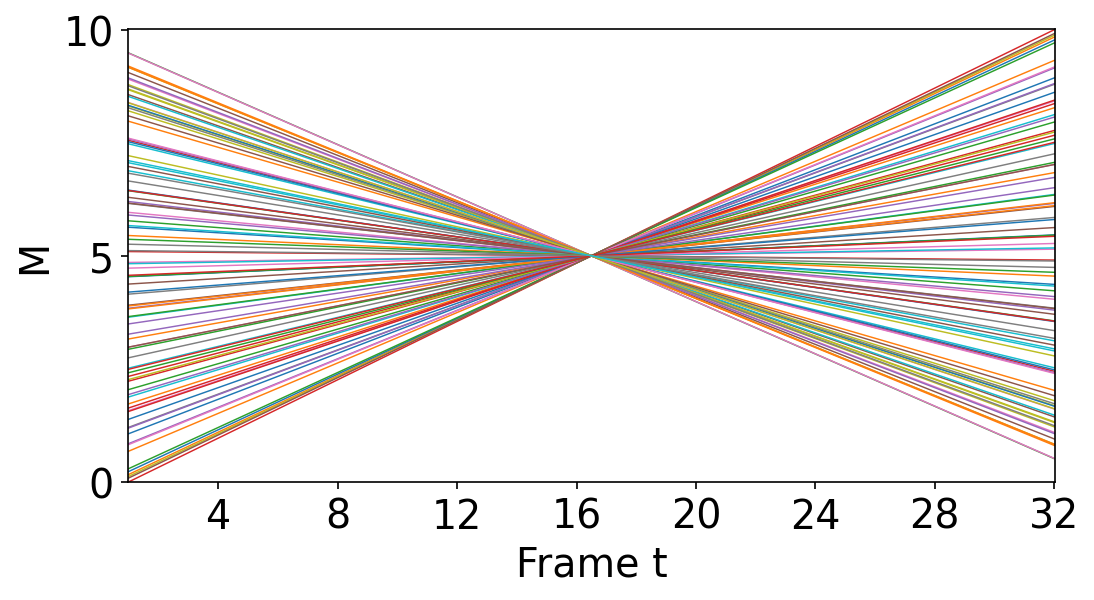}
        \caption{\small{Linear w/o Offset.}}
	\end{subfigure}
	\hfill
	\begin{subfigure}[c]{0.19\textwidth}
		\centering
        \includegraphics[width=\linewidth]{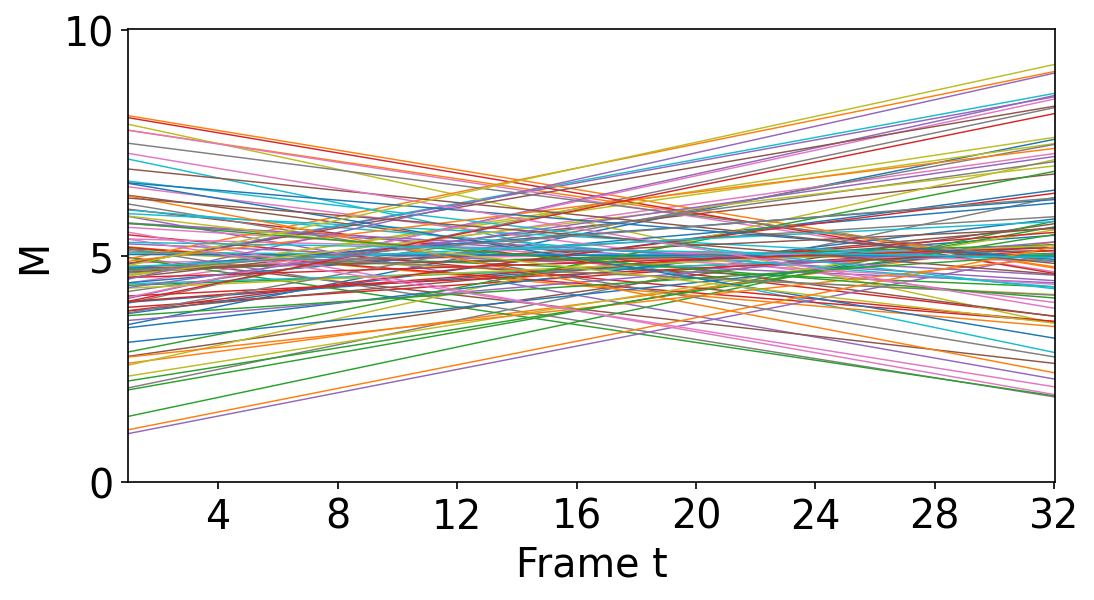}
        \caption{\small{Linear w Offset.}}
	\end{subfigure}
		\hfill
	\begin{subfigure}[c]{0.19\textwidth}
		\centering
        \includegraphics[width=\linewidth]{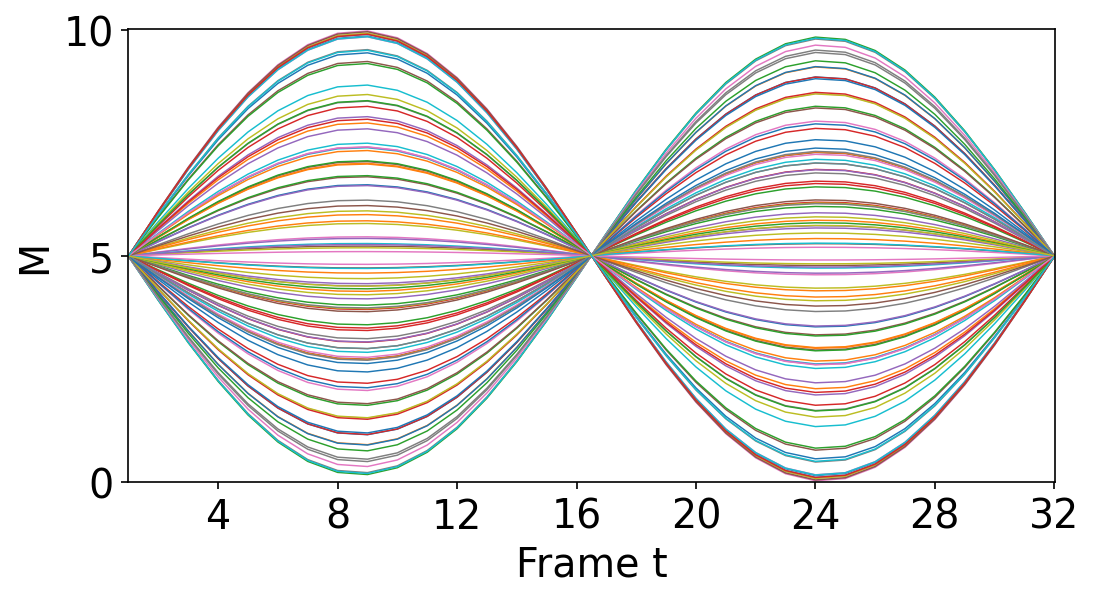}
        \caption{\small{Sine. w/o Offset.}}
	\end{subfigure}
		\hfill
	\begin{subfigure}[c]{0.19\textwidth}
		\centering
        \includegraphics[width=\linewidth]{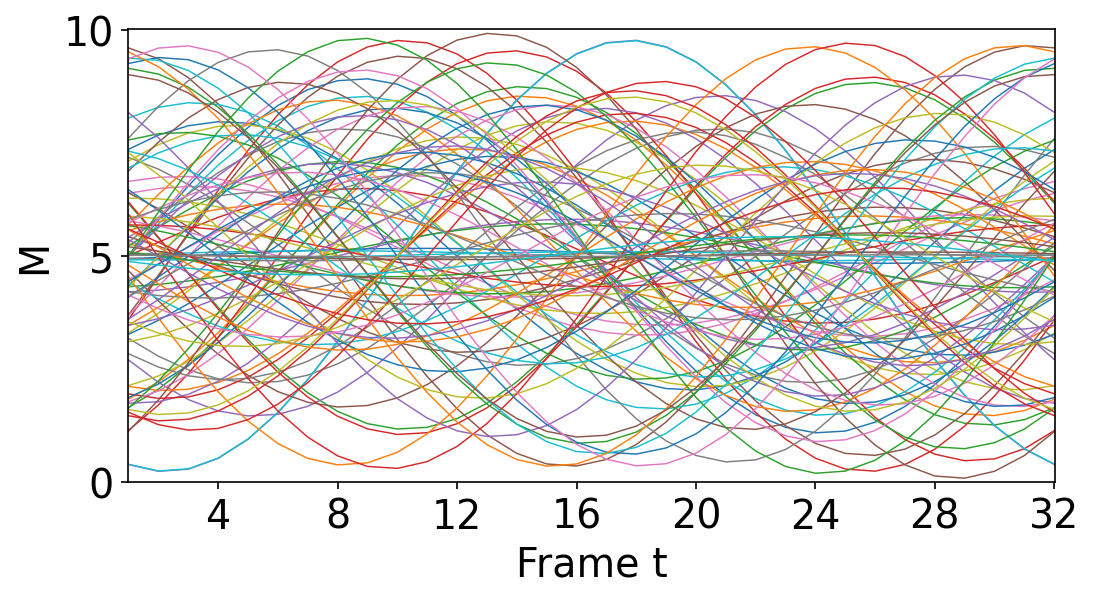}
        \caption{\small{Sine w Offset.}}
	\end{subfigure}
		\hfill
	\begin{subfigure}[c]{0.19\textwidth}
		\centering
        \includegraphics[width=\linewidth]{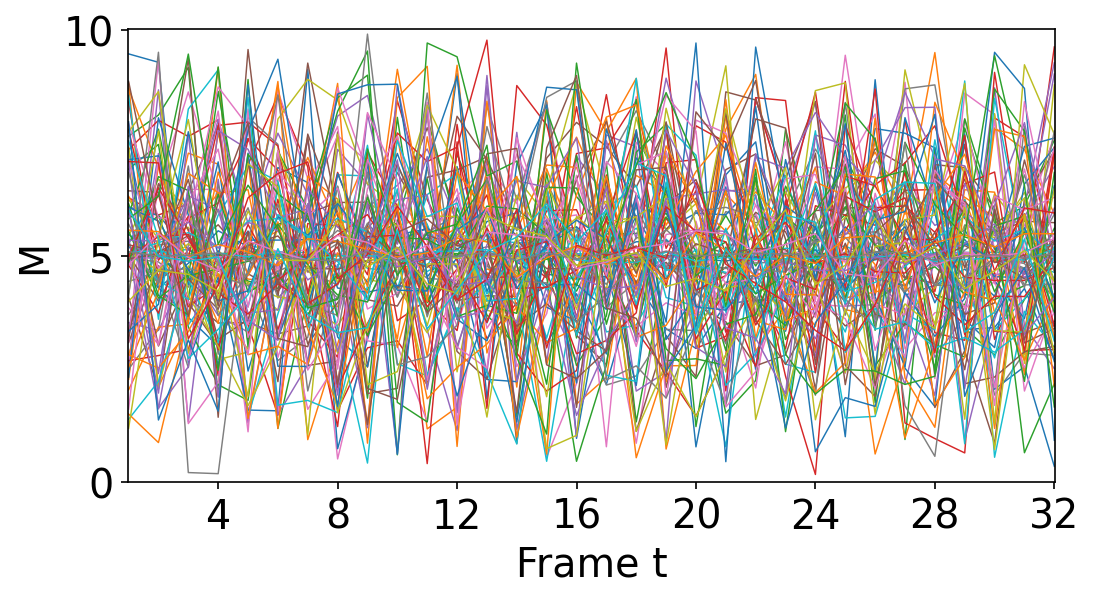}
        \caption{\small{Random}}
	\end{subfigure}
	\\
\caption{Visualizations for Table 6.}
	\label{fig:smoothness}
\end{figure}

In Table 6 (b), we compare DynaAugment's temporal variation (Fourier Sampling) with the three baselines: Linear, Sinusoidal, and Random.
We visualize the three baselines in Fig.~\ref{fig:smoothness}.
In this figure, we set the $M=5.0$ and use 32 frames for the visualization.

For Linear, we use the same policy for the amplitude and the offset with the Fourier Sampling (Fig.~\ref{fig:smoothness} (b)).
For Sinusoidal, from the Fourier Sampling, we set the range of random frequency to scalar 1.0, and we use the number of basis $C=1$ (Fig.~\ref{fig:smoothness} (d)).
For Random, a magnitude for each frame is randomly sampled as in Fig.~\ref{fig:smoothness} (e).

\subsection{Details on the Affinity and the Diversity}

% How to Calculate Affinity and Diversity, and what is the meaning?
In Table 6 (c), we use \textit{the affinity} and \textit{the diversity} as a evaluation metric for data augmentation methods. These metrics are introduced in~\cite{gontijo2020tradeoffs}.
\textit{The affinity} is calculated by the ratio between the validation accuracy of a model trained on clean data and tested on an augmented validation set (including stochastic operations), and the accuracy of the same model tested on clean data.
\textit{The diversity} is calculated by the the ratio of the final training loss of a
model trained with a given augmentation (also, including randomness), relative to the final training loss of the model trained on clean data.
The affinity is a metric for distribution shift and the diversity is a measure of augmentation complexity.
\cite{gontijo2020tradeoffs} find that performance is only improved when an augmentation operation increases the total number of unique training examples. 
The utility of these new training examples is informed by the augmentation’s (high) affinity and (high) diversity.

\subsection{Details on the Corrupted Videos}

Corrupted Kinetics validation set used in Fig. 4 is generated from the original validation set of Mini-Kinetics-200~\cite{xie2018rethinking} using following command: \texttt{ffmpeg -i [target\_video] -c:v libx264 -preset slow -crf [QP] .} where ffmpeg library~\cite{tomar2006converting} is used, \texttt{target\_video} is a directory of original validation video, and \texttt{QP} is a quantization parameter of H.264~\cite{wiegand2003overview}.
RAFT~\cite{teed2020raft} is used for optical flow visualization in Fig. 4 (b).

\section{Additional Results}

\subsection{Confidences in Results}

We run all the experiments three times unless specified, and all results in the manuscript are averaged over three runs.
For all recognition results, the standard deviations of Top-1 accuracy are less than 0.19\%.
Most of cases, because DA's improvement gaps over static augmentations are greater than 0.19\% except in some cases (\textit{e.g.}, TDN-RA in K400), the superiority of DA is mostly guaranteed.

For transfer learning results, we describe standard deviations in Section B.4.
We run all the experiments five times for the transfer learning tasks.

\subsection{More Baseline Results}

In Table~\ref{tab:more_baselines}, we show the results of more baseline for RA~\cite{cubuk2020randaug}, TA~\cite{muller2021trivialaugment}, and UA~\cite{lingchen2020uniformaugment} in Kinetics-100.
The results show that our baseline choice is reasonable because all these baselines (marked as underline and bold) show the best performances among the same augmentation methods.
Especially, in RA, the performance reduces in the other hyper-parameters.

\begin{table}[!h]
    \caption{More Baseline Results on Kinetics-100: For, RA (N, M), N is the number of operations, and M is the magnitude. E means the extended search space. For RA, default probability is 1.0. \underline{Underline} setting is the second best baseline, and \textbf{Bold} setting is the best baseline among the same augmentation method.}
    \label{tab:more_baselines}
    \centering
    \small
    \resizebox{0.9\textwidth}{!}{
        \begin{tabular}{lc|lc|lc}
            \toprule
            Config. & Top-1/Top-5 & Config. & Top-1/Top-5 & Config. & Top-1/Top-5 \\
            \midrule
            Baseline & 66.6 / 85.5 & & & & \\
            \midrule
            \underline{RA (2, 9)} &	72.2 / 90.0	& TA &	71.1 / 88.4	 & UA &	70.3 / 88.1 \\
            \textbf{RA-E (2, 9)} &	73.0 / 90.1 &	\underline{TA-Wide}	& 72.3 / 89.4 &	\underline{UA-Wide}	 & 71.9 / 89.3 \\
            RA-E (1, 9) &	71.8 / 89.1	 & \textbf{TA-E}	& 72.7 / 89.9 &	\textbf{UA-E}	& 72.8 / 89.7 \\
            RA-E (2, 7) &	71.1 / 89.0	 & & & & \\
            RA-E (3, 9) &	71.7 / 90.0	& & & & \\
            RA-E (4, 9) &	70.6 / 89.4	 & & & & \\
            RA-E (4, 7) $p=0.5$ &	71.7 / 88.9	& & & & \\
            RA (2, 9) Wide &	67.8 / 88.7	& & & &  \\
            \midrule
            \underline{RA (2, 9) + DA}	& 73.4 / 90.2 &	\underline{TA + DA}	& 73.2 / 90.0 &	\underline{UA + DA}	& 72.9 / 90.0 \\
            \textbf{RA-E (2, 9) + DA}	& 73.6 / 90.8 &	\textbf{TA-E + DA} &	73.7 / 90.5 &	\textbf{UA-E + DA} &	73.7 / 90.3 \\
            \bottomrule
        \end{tabular}}
\end{table}

In Table~\ref{tab:more_comparisons}, we compare DA results with more augmentation methods in Something-Something-v2 and UCF-101 datasets, such as, RMS~\cite{kim2020regularization}, MixUp~\cite{zhang2017mixup}, VideoMix~\cite{yun2020videomix}, and AutoAugment (AA)~\cite{cubuk2019autoaugment}. 
For RMS, We use mean filter for the pooling operation, and random sampling distribution is Gaussian whose mean is $1.0$ and standard deviation is $0.5$. RMS operation is inserted before the 3rd batch normalization layer of all residual blocks.
The mixing ratio between two data samples is sampled from $\texttt{Beta}(1.0, 1.0)$ that is identical to  $\texttt{Uniform}(0.0, 1.0)$.
The mixing is occurred within the mini-batch.
For VideoMix, the box coordinates are uniformly sampled following the original implementation.
For video, MixUp and VideoMix are applied identically across frames.
In the case of CutMix~\cite{yun2019cutmix}, it is identical to the spatial version of VideoMix.
For AA, we use searched policy from the ImageNet as in RA.

\begin{table}[!h]
    \caption{Comparison with more baselines: RMS~\cite{kim2020regularization}, MixUp~\cite{zhang2017mixup}, VideoMix~\cite{yun2020videomix}, and AutoAugment (AA)~\cite{cubuk2019autoaugment}, on Something-Something-v2, UCF-101, and Diving-48 datasets. We report Top-1 and Top-5 accuracies.}
    \label{tab:more_comparisons}
    \centering
    \small
    \resizebox{0.7\textwidth}{!}{
        \begin{tabular}{l|cc|c|c}
            \toprule
            Dataset & \multicolumn{2}{c|}{Something-v2} & UCF-101 & Diving-48 \\
            \midrule
            Model & SlowFast8$\times$8 & SlowFast16$\times$8 & S3D-G & S3D-G \\
            \midrule
            Baseline &	61.4 / 85.8 &	63.0 / 88.5 &	72.0 / 90.6 &	62.8 / 93.6 \\
            RA &	63.1 / 87.3 &	64.4 / 88.7 &	82.8 / 96.4 &	63.6 / 93.8 \\
            RA+DA &	\textbf{65.0 / 89.0} &	65.5 / 89.5 &	\textbf{85.3 / 97.2} &	70.5 / 95.6 \\
            TA &	63.5 / 89.0 &	64.7 / 88.6 &	81.6 / 96.3 &	68.3 / 95.1 \\
            TA+DA &	64.8 / 88.9 &	\textbf{65.8 / 89.8} &	82.8 / 95.7 &	\textbf{72.5 / 95.7} \\
            UA &	63.1 / 88.7 &	64.6 / 88.5 &	82.5 / 96.1 &	67.8 / 95.0 \\
            UA+DA &	64.0 / 88.9	 & 65.7 / 89.6 &	82.9 / 95.6 &	70.0 / 94.9 \\
            \midrule
            RMS~\cite{kim2020regularization} &	62.1 / 86.8 &	63.9 / 88.0 &	-  &	- \\
            MixUp~\cite{zhang2017mixup} &	62.4 / 87.3 &	63.5 / 88.4 &	76.7 / 91.8 &	\textbf{65.6 / 93.8} \\
            VideoMix~\cite{yun2020videomix} &	\textbf{63.2 / 88.2} &	64.2 / 88.4 &	73.8 / 91.9 &	63.0 / 93.4 \\
            AA~\cite{cubuk2019autoaugment} &	63.1 / 87.4 &	\textbf{64.6 / 88.6} &	\textbf{81.7 / 94.9} &	63.1 / 93.6 \\
            \bottomrule
        \end{tabular}}
\end{table}

\subsection{UCF/HMDB Results on the Pre-trained Models}

We describe the results on UCF-101~\cite{soomro2012ucf101} and HMDB-51~\cite{kuehne2011hmdb} datasets using ImageNet or Kinetics pre-training in Table~\ref{tab:ucf_full}.
From the results, we can check the effects of augmentations in terms of model pre-training.
Although the improvement of performance are reduced compare to \textit{training-from-scratch} through the pre-training, DA still increases the performances.

\begin{table}[!h]
    \caption{Results on UCF-101 and HMDB-51 datasets with different pre-training settings.}
    \label{tab:ucf_full}
    \centering
    \small
    \resizebox{1.0\textwidth}{!}{
        \begin{tabular}{ll|ccccccc}
            \toprule
            Dataset & Model & Baseline & RA & RA + DA & TA & TA + DA & UA & UA + DA \\
            \midrule
            UCF-101 & S3D-G from Scratch & 72.0 / 90.6 &	82.8 / 96.4 &	\textbf{85.3 / 97.2} &	81.6 / 96.3 &	\textbf{82.8 / 95.7} &	82.5 / 96.1 &	\textbf{82.9 / 95.6} \\
             & I3D-16$\times$8 from ImageNet & 79.7 / 93.7 &	87.4 / 97.8 &	\textbf{88.7 / 98.0} &	86.2 / 97.2 &	\textbf{87.3 / 97.5} &	85.8 / 97.3	& \textbf{87.6 / 97.9} \\
             & I3D-16$\times$8 from Kinetics & 92.4 / 98.9 &	93.5 / 98.7 &	\textbf{94.5 / 98.9} &	94.4 / 99.0 &	\textbf{94.5 / 99.1} &	94.0 / 99.2 &	\textbf{94.8 / 99.3} \\
             \midrule
             HMDB-51 & S3D-G from Scratch & 40.1 / 72.6	 & 48.5 / 78.4 &	\textbf{54.8 / 83.7} &	47.7 / 78.9 &	\textbf{51.9 / 80.7} &	49.9 / 79.8 &	\textbf{52.4 / 78.9} \\
             & I3D-16$\times$8 from ImageNet & 44.9 / 72.3 &	53.6 / 80.9 &	\textbf{54.4 / 83.8} &	51.6 / 79.7 &	\textbf{52.7 / 81.4} &	50.9 / 79.6	& \textbf{51.2 / 79.9} \\
             & I3D-16$\times$8 from Kinetics & 62.9 / 86.0 &	64.9 / 88.5 &	\textbf{66.8 / 88.9} &	64.7 / 87.5 &	\textbf{67.0 / 89.0} &	67.0 / 89.9	 & \textbf{67.5 / 89.9} \\
            \bottomrule
        \end{tabular}}
\end{table}

\subsection{More Results on Transfer Learning}

In Table~\ref{tab:transfer_full}, we describe TA results for Breakfast~\cite{Kuehne12breakfast} dataset and UA results for MOT17det~\cite{milan2016mot16} dataset with their standard deviations. These results show the similar trend as in the main manuscript.

\begin{table}[t!]
	\centering
	\caption{Full results with standard deviations of Table 5.}\label{tab:transfer_full}
	
\begin{subtable}[t]{.5\textwidth}
\centering
\caption{Action segmentation results on Breakfast dataset.}\label{tab:seg_appendix}
\setlength{\tabcolsep}{3pt}
\small
\vspace{-0.5mm}

\small
\begin{tabular}{lccccc}
    \toprule
        &   & \multicolumn{3}{c}{F1}  \\
    \cmidrule{3-5}
    \textbf{Configuration} & Acc. & @0.10 & @0.25 & @0.50 \\
    \midrule
    SO-50 & 59.0 & {{54.7}\ms{2.0}\hfill} & {{49.2}\ms{1.7}\hfill} & {{37.6}\ms{1.8}\hfill} \\ % 2.0 / 1.7 / 1.8
    SO-50+TA & {62.0} & {{58.8}\ms{1.2}\hfill} & {{52.7}\ms{1.4}\hfill} & {{40.2}\ms{1.6}\hfill} \\
    SO-50+TA+\textbf{DA} & \textbf{64.3} & \textbf{{{60.1}\ms{1.2}\hfill}} & \textbf{{{54.9}\ms{1.6}\hfill}} & \textbf{{{42.8}\ms{1.5}\hfill}} \\
    SO-50+UA & {62.3} & {{59.1}\ms{1.1}\hfill} & {{53.7}\ms{1.6}\hfill} & {{40.8}\ms{1.4}\hfill} \\
    SO-50+UA+\textbf{DA} & \textbf{64.7} & \textbf{{{60.6}\ms{1.0}\hfill}} & \textbf{{{55.5}\ms{1.4}\hfill}} & \textbf{{{43.2}\ms{1.5}\hfill}} \\
    \bottomrule
\end{tabular}

\end{subtable}
\begin{subtable}[t]{.49\textwidth}

\caption{Object detection results on MOT17det dataset.}\label{tab:det_appendix}
\begin{center}
\small
\vspace{-2.3mm}
\setlength{\tabcolsep}{4pt}

\small
\begin{tabular}{lccccc}
    \toprule
    \textbf{Configuration} & AP & AP50 & AP75 & APs \\
    \midrule
    Swin-T & {{30.3}\ms{0.9}\hfill} & {{63.7}\ms{1.3}\hfill} & {{25.1}\ms{1.4}\hfill} & {{4.1}\ms{0.8}\hfill} \\
    Swin-T+TA & {{29.1}\ms{1.3}\hfill} & {{62.1}\ms{1.6}\hfill} & {{23.2}\ms{2.3}\hfill} & {{3.7}\ms{0.7}\hfill} \\
    Swin-T+TA+\textbf{DA} & \textbf{{{31.3}\ms{0.6}\hfill}} & \textbf{{{65.3\ms{0.8}\hfill}}} & \textbf{{{26.3\ms{1.0}\hfill}}} & \textbf{{{5.7\ms{0.3}\hfill}}} \\
    Swin-T+UA & {{29.4}\ms{1.1}\hfill} & {{62.9}\ms{1.5}\hfill} & {{23.4}\ms{1.6}\hfill} & {{4.1}\ms{1.5}\hfill} \\
    Swin-T+UA+\textbf{DA} & \textbf{{{31.4}\ms{0.7}\hfill}} & \textbf{{{65.5\ms{0.5}\hfill}}} & \textbf{{{25.9\ms{1.8}\hfill}}} & \textbf{{{4.4\ms{0.1}\hfill}}} \\
    \bottomrule
\end{tabular}
\end{center}
\end{subtable}
\end{table}

\subsection{More Results on Fourier Sampling}

In Table~\ref{tab:fourier_full}, we describe more ablation studies on Fourier Sampling as explained in Section A. 4. in UCF-101 dataset.
For the number of basis, any choice shows superiority over the baseline and the static version.
For the amplitude, a wide range shows better and more stable results.
For the range of frequencies, because the other setting also shows similar performances, we choose the range by the heuristic.
Adding random offset can generate more diverse signals, so the performance is increased.
Gaussian smoothing over random variations can be a substitute for Fourier Sampling. However, it is not only inexplicable but also difficult to improve performance when implemented simply.

\begin{table}[!h]
    \caption{More ablation studies on Fourier Sampling. We describe Top-1 and Top-5 accuracies on UCF-101 dataset. \textbf{Bold} indicates our default setting.}
    \label{tab:fourier_full}
    \centering
    \small
    \resizebox{0.65\textwidth}{!}{
        \begin{tabular}{lc|lc}
            \toprule
            Config. & Acc. & Config. & Acc. \\
            \midrule
            Baseline &	72.0 / 90.6	 & & \\
            $C=1$ & 	84.4 / 97.1	 & & \\
            $C=2$ &	84.7 / 97.0 & 	$f_{b}$=[0.2 $\sim$ 1.5] &	\textbf{85.3 / 97.2} \\
            \textbf{$C=3$} &	\textbf{85.3 / 97.2} &	\textbf{$f_{b}$=[0.2 $\sim$ 1.0]} &	85.1 / 97.0 \\
            $C=4$ &	85.2 / 97.1 &	$f_{b}$=[0.5 $\sim$ 2.0] &	84.6 / 96.9 \\
            $A$=[0.0 $\sim$ 0.5] &	83.5 / 96.8 &	\textbf{w Offset} &	\textbf{85.3 / 97.2} \\
            $A$=[0.5 $\sim$ 1.0] &	83.9 / 97.0 &	w/o Offset &	84.7 / 96.5 \\
            \textbf{$A$=[0.0 $\sim$ 1.0]} &	\textbf{85.3 / 97.2} &	Random+Gaussian	 & 83.1 / 96.4 \\
            \bottomrule
        \end{tabular}}
\end{table}

\section{Social Impacts}

We propose a generalized algorithm for training deep networks.
Therefore, our algorithm may not exert direct negative social impacts.
Because our method has improved the performance of the models, we expect to reduce the effort and cost of other efforts to improve the performance, such as hyper-parameter search or architecture search.
However, on the one hand, our method also contains new hyper-parameters, so it is expected to be expensive to find a better optimal setting.
The search for hyper-parameter tuning can result in a lot of carbon footprint and computational cost, which can be a potential negative social impact.

%%%%%%%%%%%%%%%%%%%%%%%%%%%%%%%%%%%%%%%%%%%%%%%%%%%%%%%%%%%%

\end{document}